\documentclass{IJCAS}

\usepackage{array,tabularx}
\usepackage{multicol}

\usepackage{cite}
\usepackage{amsmath,amssymb,amsfonts}
\usepackage{algorithmic}
\usepackage{graphicx}
\usepackage{textcomp}
\usepackage{paralist}
\usepackage{mathcomp}
\usepackage{subfigure} 
\usepackage{color}
\usepackage{here}
\newcolumntype{P}[1]{>{\centering\arraybackslash}p{#1}}
\usepackage{booktabs}
\usepackage{multirow}
\usepackage[table,xcdraw]{xcolor}
\usepackage{url}
\usepackage{enumitem}
\usepackage{makecell}
\usepackage{anyfontsize}
\usepackage{adjustbox}
\usepackage{hyperref}
\usepackage[lined,boxed,linesnumbered,ruled]{algorithm2e}

\journalvolumn{VV}
\journalnumber{X}
\journalyear{YYYY}
\setarticlestartpagenumber{1}
\definecolor{ForestGreen}{RGB}{34, 139, 34}

\allowdisplaybreaks

\begin{document}
%% TITLE, AUTHOR, ABSTRACT, KEYWORDS

% the \title command
\title{Vision-based Autonomous Driving for Unstructured Environments \\ Using Imitation Learning}

% the \author command
\author{Joonwoo Ahn\orcid{0000-0001-6217-1356}, Minsoo Kim\orcid{0000-0003-2300-1279}, and Jaeheung Park*\orcid{0000-0002-5062-8264}
}%  <- don't remove this stop

% the abstract environment
\begin{abstract}
Unstructured environments are difficult for autonomous driving.
This is because various unknown obstacles are lied in drivable space without lanes, and its width and curvature change widely.
In such complex environments, searching for a path in real-time is difficult. Also, inaccurate localization data reduce the path tracking accuracy, increasing the risk of collision. Instead of searching and tracking the path, an alternative approach has been proposed that reactively avoids obstacles in real-time. 
Some methods are available for tracking global path while avoiding obstacles using the candidate paths and the artificial potential field.
However, these methods require heuristics to find specific parameters for handling various complex environments.
In addition, it is difficult to track the global path accurately in practice because of inaccurate localization data.
If the drivable space is not accurately recognized (i.e., noisy state), the vehicle may not smoothly drive or may collide with obstacles. In this study, a method in which the vehicle drives toward drivable space only using a vision-based occupancy grid map is proposed. The proposed method uses imitation learning, where a deep neural network is trained with expert driving data. The network can learn driving patterns suited for various complex and noisy situations because these situations are contained in the training data. 
Experiments with a vehicle in actual parking lots demonstrated the limitations of general model-based methods and the effectiveness of the proposed imitation learning method.
\end{abstract}

% the keywords environment

\maketitle

\makeAuthorInformation{
% Manuscript received January 10, 2015; revised March 10, 2015; accepted May 10, 2015. Recommended by Associate Editor Soon-Shin Lee under the direction of Editor Milton John. \\

Joonwoo Ahn, Minsoo Kim, and Jaeheung Park are with Dynamic Robotics Systems (DYROS) Lab., 306, builing 18, Seoul National University, 1, Gwanak-ro, Gwanak-gu, Seoul, Republic of Korea (e-mail: \{joonwooahn, msk930512, park73\}@snu.ac.kr). \\
Jaeheung Park is also in Advanced Institutes of Convergence Technology (AICT), Suwon-si, Republic of Korea.
* Corresponding author. \\
This work has been submitted to the Springer for possible publication.
Copyright may be transferred without notice, after which this versoin may no longer be accessible.
}

% \runningtitle{2020}{Joonwoo Ahn, Minsoo Kim, and Jaeheung Park}{Manuscript Template for the International Journal of Control, Automation, and Systems: ICROS {\&} KIEE}{xxx}{xxxx}{x}

\section{Introduction}
\label{sec:intro}
Autonomous driving technology for unstructured environments such as parking lots and alleyways is important to realize fully autonomous driving.
Also, it is more difficult than driving in structured environments.
% This is because it is necessary for fully autonomous driving and is more difficult than driving in structured environments.
In a structured environment, autonomous driving involves a global plan with a road network, and a vehicle stays within a lane through lateral control and maintains a safe distance from vehicles in front while following a target speed through longitudinal control. 
% In the case of driving in structured environments, a global plan with a road network is conducted, and a vehicle keeps the lane on roads through lateral control and maintains a distance from the front vehicle while following a target speed by longitudinal control. 
However, applying this method to unstructured environments is difficult as drivable space has no lanes and variable width.
% However, this method is difficult to be applied to unstructured environments.
% It is because there are no lanes on the drivable space of these environments, and the width of the drivable space is various.
Moreover, the curvature can rapidly change, such as at right-angled corners, and the drivable space can be narrowed because of double-parking or illegal parking.
% Moreover, the curvature changes rapidly, such as the right-angled corner, and the drivable space can be narrowed due to double parking or illegal parking.
Other obstacles include vehicles, humans, curbs, and bollards, which vary in shape, size, and location. 
% Besides, there are obstacles such as vehicles, humans, kerbs, and bollards, which vary in shape, size, and location. 
Such obstacles typically are unknown in advance.
% These typically may not be known in advance.

\begin{figure*}[t]
\centering
	\includegraphics[width = 1.0\linewidth]{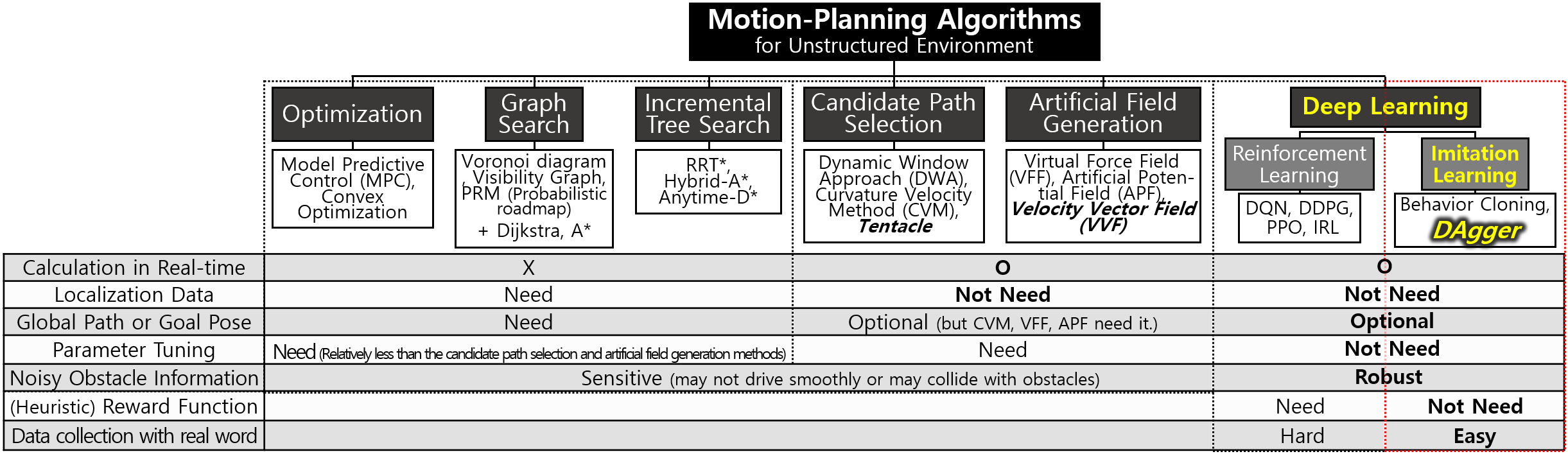}
    \caption{
        Motion-planning algorithms and comparison between the characteristics of sub-algorithms.
        % motion-planning algorithms and characteristics comparison between sub-algorithms
    }
    \label{fig:motion_planning}
\end{figure*}

For a vehicle to drive in an unstructured environment, a global path is generated on a global map to reach the destination.
% To drive in unstructured environments, a global route is generated through a global map, which can reach the destination.
The vehicle tracks the path based on localization data (i.e., the position and heading of the vehicle relative to the path) \cite{banzhaf2017future}.
% A vehicle tracks it with localization data that is the position and heading of the vehicle relative to the path \cite{banzhaf2017future}.
While tracking the global path, the vehicle checks for obstacles in its vicinity.
% While tracking the global path, the vehicle checks for obstacles around the vehicle.
Object detection algorithms detect the position and size of obstacles by using camera or LiDAR sensors with pattern recognition or deep learning.
% The object detection algorithm detects the position and size of the obstacles through the camera or LiDAR sensors, which is based on pattern recognition or deep learning.
If obstacles are detected near the global path, motion-planning is performed to find a local path or waypoint that can reach the global path without collision. % by localization data
% If it is determined that detected obstacles are near the global path, motion-planning is performed to find a local path or a waypoint that can reach the global path without collision.
The planned solution must also satisfy dynamic and kinematic constraints on the motion of the vehicle.
% These also satisfy dynamic and kinematic constraints on the motion of the vehicle.
Motion-planning algorithms developed for robotics have been applied to autonomous vehicles \cite{paden2016survey, gonzalez2015review}.
% motion-planning algorithms discussed in the robotics literature have been applied to an autonomous vehicle \cite{paden2016survey, de1997computational, gonzalez2015review}. %
These can be categorized according to the method and calculation time.
% These can be categorized according to the method of finding a solution and the calculation time.
Figure \ref{fig:motion_planning} shows an overview of motion-planning algorithms, which include optimization, graph search, and incremental search path planning methods to find a solution for the local area.
% An overview of motion-planning is shown in Fig. \ref{fig:motion_planning}.
% 
% There are optimization, graph search, and incremental search path planning methods to find a solution for the entire local area.

The path planning method using optimization theory, such as model predictive control (MPC) \cite{shin2021model} and convex optimization \cite{mousavi2017new}, uses a kinematic and dynamic model of the vehicle to predict its future trajectory.
% The path planning method using the optimization theory, such as model predictive control (MPC) \cite{shin2021model} and convex optimization \cite{mousavi2017new} uses the kinematic and dynamic model of the vehicle, which can predict a future trajectory of the vehicle.
This method provides an optimal solution that satisfies the objective function and constraints.
% This method provides an optimal solution that satisfies an objective function and constraints.
In driving situations, the objective function can be modeled as avoiding obstacles while reaching the global path and maintaining target speed.
% In driving situations, the objective function can be modeled to avoid obstacles while reaching the goal point and maintaining target speed.
Constraints can be the control capabilities and maintaining a safe distance from obstacles.
% The constraint can be a control capability and a safe distance from the obstacles.

The graph search path planning method builds a graph in the local area and then searches for a path.
% The graph search path planning method builds a graph in the local area and then searches a path.
The Voronoi diagram \cite{al2020voronoi}, Visibility graph \cite{niu2019voronoi}, and Probabilistic roadmap (PRM) \cite{mohanta2019knowledge} algorithms can be used to build the graph.
% The voronoi diagram \cite{garrido2006path}, visibility graph \cite{niu2019voronoi}, probabilistic roadmap (PRM) \cite{mohanta2019knowledge} algorithms are used to build the graph.
These algorithms discretize the configuration space into obstacles and free space, which are represented in the form of a graph.
% These algorithms discretize the configuration space consisting of obstacles and free space and represent it in the form of a graph.
Then, the graph is searched for the minimum path length with the Dijkstra \cite{dijkstra1959note} or A* \cite{hart1968formal} graph search algorithm.
% Then, a path with minimum length is searched in the graph using the Dijkstra \cite{dijkstra1959note} or A* \cite{hart1968formal} graph search algorithms.
The searched path is interpolated through spline algorithms to satisfy vehicle constraints and obtain a smooth path.
% The searched path is interpolated through the spline algorithms to satisfy the constraint of the vehicle and obtain a smooth path.

The incremental search path planning method uses tree exploration algorithms.
These algorithms iteratively expand a tree into free space until the end of the tree reaches a goal.
% These algorithms iteratively expand a tree to the point that there are no obstacles until to reaching a goal pose.
% These algorithms expand a tree to a point where there are no obstacles until to reach a goal pose.
The rapidly-exploring random trees$^*$ (RRT$^*$) algorithm \cite{shin2016desired} extends the tree with samples randomly selected in the configuration space.
The hybrid-A$^*$ \cite{dolgov2009path} and anytime-D$^*$ \cite{likhachev2008anytime} algorithms expand the tree in grid units.
Then, the path with the minimum length is searched for to reach the goal pose while satisfying the non-holonomic constraints of the vehicle.
% The searched path has a minimum length to reach the goal pose and can satisfy the non-holonomic constraint of the vehicle.

However, these methods have three problems \cite{hoy2015algorithms}.
First, if the local area is large or complex, a long computational time is needed to generate the path, and the solution may not be found within a control loop (i.e., not real-time).
% First, if the local area is large or complex, it takes a long computational time to generate the path and may not able to find the solution within a control loop (not real-time).
Second, selecting a goal in the global path to search for the local path is heuristic.
% Second, a method for selecting a goal in the global path to search the local path is not clear.
Third, when an algorithm is implemented, accurately recognizing whether an obstacle is near the global path and tracking the path without collision are difficult because of inaccurate localization data.
% Third, when implementing the algorithms, it is difficult to accurately recognize whether an obstacle exists near the global path and to track the path without collision due to inaccurate localization data.
% In unstructured environments, there are more cases with a narrow drivable space and obstacles with diverse and complex shapes and sizes than in structured environments.
In unstructured environments, various types of obstacles are complexly placed in the drivable area.
% In unstructured environments, there are more cases where the drivable space is narrow and the shape and size of obstacles are diverse and complex than in the structured environment.
Thus, obtaining accurate localization data at every point in unstructured environments is difficult.
% Thus, it is difficult to obtain accurate localization data at every point in unstructured environments.

Instead of searching and tracking a path, alternative methods can be used, including candidate path selection and the artificial field generation \cite{hoy2015algorithms}.
These methods find a solution near a vehicle that can be calculated in real-time.
They select a candidate path or waypoint and calculate the control commands.
The vehicle can then drive toward the global path while reactively avoiding obstacles.

The candidate path selection method generates candidate paths and selects one path that satisfies multiple objectives.
These paths are smooth and are designed to account for the non-holonomic constraints of the vehicle.
% These paths are smooth and designed to take into account the non-holonomic constraint of the vehicle.
To select one path, the objective function is modeled to reach the global path, avoid obstacles, and keep a ride comfort.
%%%% To select one path, the objective function is modeled to maximize the distance between the vehicle and obstacles as well as minimize changes in the steering angle and velocity.
% This function also includes terms that can reach the global path.
% Each term is applied with a different weight.
Typically, three algorithms have been used:
% Typically, there are three algorithms according to the way of modeling the candidate paths; 
the Dynamic window approach (DWA) \cite{missura2019predictive}, Curvature velocity method (CVM) \cite{lopez2019new}, and tentacle \cite{mouhagir2019evidential} algorithms.
The DWA algorithm designs a window according to the current state of the vehicle, and candidate paths are generated within the window.
The CVM algorithm is similar to DWA and additionally considers accelerations of the vehicle.
% The CVM algorithm is similar to DWA and additionally takes into account the accelerations of the vehicle.
The tentacle algorithm mimics the antennas of a beetle as candidate paths to drive on narrow and variable-curvature roads more smoothly than DWA and CVM.
% It can drive on narrow and varied curvature roads more smoothly than DWA and CVM.

The artificial field generation method uses a repulsive field against obstacles and attractive field toward the global path.
% The artificial field generation method has a repulsive field against obstacles and an attractive field toward the global path.
These fields are combined with different weights, and a vehicle is guided by the combined field's vector.
Three algorithms are available that differ on how to model the fields:
% Depending on the way of modeling the fields, there are three algorithms; 
the virtual force field (VFF) \cite{olunloyo2009autonomous}, artificial potential field (APF) \cite{ge2000new}, and velocity vector field (VVF) \cite{wang2015driving} algorithms.
The VFF algorithm calculates the repulsive force as a vector from the obstacle to the vehicle and the attractive force as a vector from the vehicle to the target point.
The APF algorithm creates a repulsive field with high potential energy for obstacles and an attractive field with high energy at the vehicle point and low energy at the goal point.
Then, the field is generated from the gradient of the potential energy.
% Then, the field is generated by the gradient of the potential energy.
The VVF algorithm considers the desired velocity and velocity of obstacles in addition to the fields of the APF algorithm.
% In the VVF algorithm, the desired velocity and the velocity of obstacles are additionally reflected in the field of APF.

However, the candidate path selection and artificial field generation methods have several problems that make them difficult to be used in unstructured environments \cite{chen2019deep, hawke2020urban}.
% However, the candidate path selection and artificial field generation methods are hard to apply in unstructured environments due to the following problems \cite{chen2019deep, hawke2020urban}.
First, the parameters (weights) in the objective function or field model may differ to cope with the various complex situations of unstructured environments.
% First, parameters (weights) in the objective function or the field model to cope with various complex situations in unstructured environments may be different.
It is not easy to find specific parameters that can handle all of these situations.
% So it is not easy to find specific parameters that can handle all these situations.
Second, inaccurate localization data make it difficult in practice to know where exactly the global path is located in a local area \cite{youn2021collision}.
% Second, in practice, it is difficult to know exactly where the global path is located in the local area due to inaccurate localization data.
Third, if the local obstacle information is difficult to recognize accurately especially at road boundaries or shadowed areas (i.e., noisy state), the vehicle may not drive smoothly \cite{varma2019idd}.
% Third, above all, if the local obstacle information is not recognized accurately (noisy) especially at road boundaries or shadow areas \cite{varma2019idd}, the vehicle may not drive smoothly.
In addition, a vehicle may drive out of the drivable space or toward an obstacle.
% Besides, the vehicle may drive out of the drivable space or toward the obstacle.

\subsection{Overview of Our Approach}
To address these problems, this study proposes a method of selecting a waypoint (look-ahead point) to drive toward the drivable space while avoiding obstacles in real-time without the use of global information\footnote{The proposed method does not use global information, assuming it only handles environments without intersections.} such as the global map, global path, and localization data.
% To address the mentioned problems, this study proposes a method that selects a waypoint to drive toward the drivable space while avoiding obstacles in real-time without using global information (global map, global path, localization data)\footnote{The proposed method does not use global information, assuming it only handles environments without intersections.}.
The proposed method only segments the drivable space and non-drivable space around a vehicle; it does not recognize whether obstacles exist near the global path.
% The proposed method only segments the drivable space and the non-drivable space around the vehicle, not recognizing whether obstacles exist near the global path.
The segmented space is represented as an occupancy grid map, which is obtained by deep learning to segment the image acquired from the camera.
% This map is obtained through deep learning that segments the image acquired from the camera. 
The motion-planning algorithm in the proposed method is based on deep learning \cite{lee2021deep}, which is an alternative to general motion-planning algorithms \cite{chen2019deep, hawke2019urban}.
% In addition, the motion-planning algorithm in the proposed method is based on deep learning, which can be an alternative to the conventional motion-planning algorithms \cite{chen2019deep, hawke2019urban}.
It can handle the various complicated situations that occur in unstructured environments without requiring the model parameters to be tuned, and it is robust against noisy obstacle information.
% It can handle the various complicated situations that occur in unstructured environments without tuning the model parameter and can be robust against the noisy obstacle information. 

In the motion-planning, training data are collected of state action pairs without the use of global information, and the data are used to train a deep neural network. 
In the data, states contain various situations that are difficult to handle with general motion-planning algorithms, such as large changes of the curvature and width of the drivable space, and noisy state.
% The states are labeled with actions appropriate for an objective.
% Actions appropriate to an objective are labeled to the states.
Using data, the network can learn patterns for all the information of the occupancy grid map, not just information around the candidate paths or artificial fields.
% In addition, the network learns patterns for entire information of the occupancy grid map, not just information around the candidate paths or the artificial fields.
%%% Thus, information about various and complex situations expressed by the map can be reflected in the driving pattern.
% Thus, information about the various and complex situations expressed in this map can be reflected in the driving pattern.
Depending on how a dataset is collected, there are reinforcement learning and imitation learning.
% Depending on the way of collecting the datasets, there are reinforcement learning and imitation learning.

For reinforcement learning, data are collected to maximize a reward.
% In reinforcement learning, datasets are collected to maximize a reward.
However, heuristics are required to model the reward function to achieve an objective \cite{hadfield2017inverse}.
% However, modeling the reward function to achieve an objective is heuristic \cite{hadfield2017inverse}.
In addition, the agent gathers data through trial and error with random actions, so the training takes a long time to complete.
% Besides, the agent gathers the data through trial and error with random actions, and it takes a long time to complete the training.
Therefore, training in a real environment is difficult, and most studies on reinforcement learning requires the use of simulations.
% Therefore, it is difficult to train in a real environment, and most studies are conducted on simulation.

In this study, imitation learning was used to collect data effectively.
% This study uses imitation learning to collect data effectively.
Imitation learning collects successful driving data obtained by experts \cite{bojarski2016end}.
% Imitation learning collects datasets that have been successful in driving by an expert \cite{bojarski2016end}.
These data can be used to train the deep neural network in an approach similar to supervised learning. 
% These datasets train the deep neural network, which is similar to supervised learning.
The trained network can imitate the driving patterns of expert driving.
% The trained network can imitate a driving pattern of expert driving.
Therefore, a heuristic model is not needed to evaluate an action.
% Therefore, it does not need a heuristic model to evaluate an action.
Because the data are collected without trial and error, less training time is required than for reinforcement learning, and data of real environments are easily obtained.
% Besides, since datasets are collected without trial and error, it takes less training time than reinforcement learning, and datasets for the real environment are easily obtained.

In the proposed method, imitation learning is used to select a look-ahead point with the occupancy grid map as the input. 
% A proposed imitation learning method learns to select the look-ahead point with the occupancy grid map as input.
% The vehicle reaches the look-ahead point by calculating the steering angle with 
The pure pursuit algorithm is used to calculate the steering angle to reach the look-ahead point.
and velocity according to the longitudinal distance between the look-ahead point and vehicle.
The dataset aggregation (\textit{DAgger}) algorithm \cite{menda2019ensembledagger} is used to address unsafe and near-collision situations occurred by the trained network policy.
By using the look-ahead point, \textit{DAgger} can be applied to the real autonomous vehicle.
In addition, we propose a method to train these problem situations faster and more accurately than \textit{EnsembleDAgger}.

% \color{red}
    Our contributions are summarized as follows:
    \begin{itemize} %[label=(\roman*)]
    \item A method is proposed to drive with only vision data in unstructured environments using imitation learning, which does not use high-cost HD-map and inaccurate localization data in a complex environment.
    
    \item Real-world experiments show limitations of the model-based motion-planning algorithms and effectiveness of the proposed method which is robust to sensor noise and does not need to tune model parameters to handle various and complex environments.
    
    \item As features and innovations, the vision information is converted into the occupancy grid map in order to apply imitation learning to this information.
    Besides, the look-ahead point of the pure pursuit algorithm, which has been widely used in autonomous driving, was used to train, which has a clear state action pattern relationship, and a safe driving policy is obtained.
    
    Moreover, \textit{DAgger} algorithm is introduced to further improve the performance of imitation learning, and (\textit{DAgger}) can be applied to the real autonomous vehicle by using the look-ahead point.
    \end{itemize}
% \color{black}

The rest of this paper is organized as follows.
% The content of this paper is organized as follows.
Section \ref{sec:method} explains the vision-based occupancy grid map and imitation learning-based driving policy.
The experimental setup and results are presented in Section \ref{sec:experimental_setup} and \ref{sec:experimental_result}.
In the experiment with an autonomous vehicle, the \textit{tentacle} algorithm, \textit{VVF} algorithm, and proposed method were tested in real parking lots with the same input (i.e., occupancy grid map).
The experimental results demonstrated that the vehicle was successfully driven with the proposed imitation learning method in situations where the \textit{tentacle} and \textit{VVF} algorithms encountered problems.
% Through this experiment, it shows that the autonomous vehicle using the proposed method successfully drives in situations where the Tentacle or VVF algorithms causes problems.
Section \ref{sec:conclusion} concludes this study and mentions future work.
%%%%%%%%%%%%%%%%%%%%%%%%%%%%%%%%%%%

\begin{figure*}[t]
\centering
	\includegraphics[width = 1.0\linewidth]{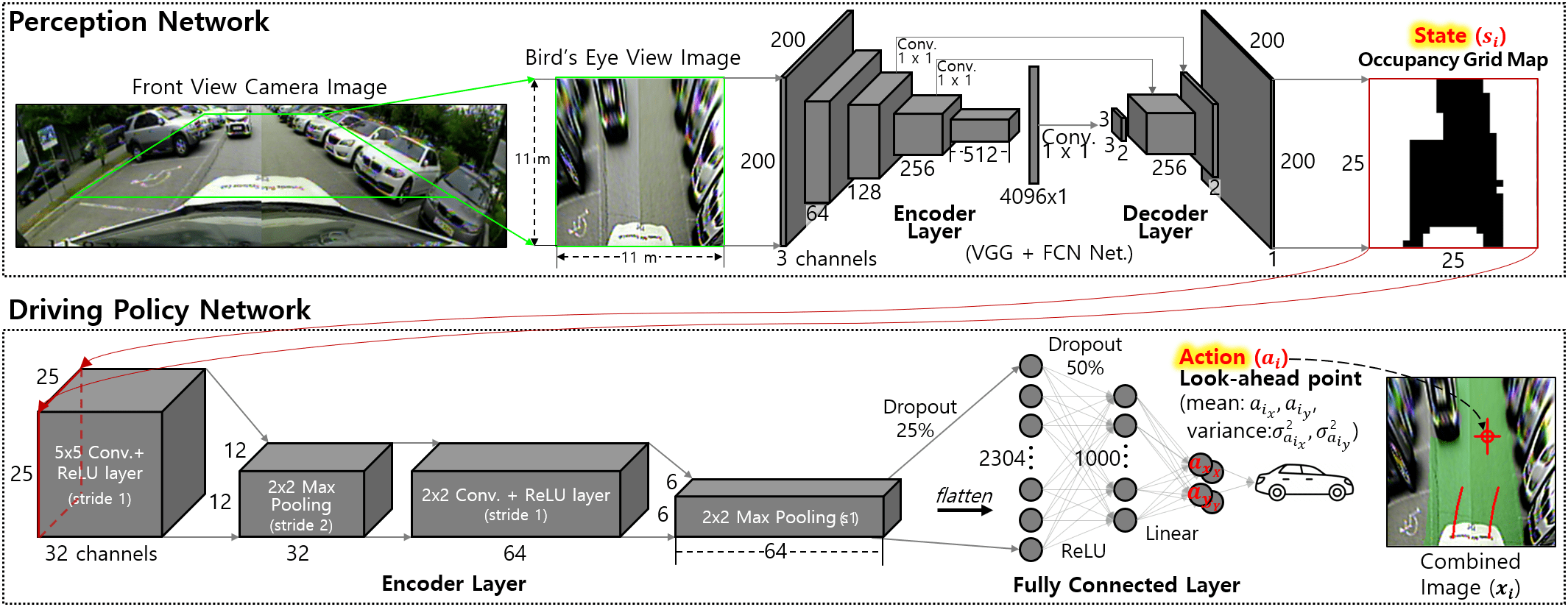}
    \caption{
        System architecture and deep neural network of the proposed methods.
    }
    \label{fig:system}
\end{figure*}

\section{Methods}
\label{sec:method}
This section presents the methods for obtaining the occupancy grid map from vision data and the driving policy in an unstructured environment through imitation learning.
The input for imitation learning is the occupancy grid map, and the output is the look-ahead point used to control the vehicle.
In this study, the road was assumed to have only static obstacles and no intersections.

\subsection{Vision-based Occupancy Grid Map}
\label{subsec:grid_map}
The occupancy grid map is a 2D map that divides an area into a grid.
% The occupancy grid map is a 2-D map that divides a certain area into a grid.
It is shown in the red box in the upper right of Fig. \ref{fig:system}.
Each grid in the map contains information on whether it is occupied (non-drivable) or unoccupied (drivable).
% In each cell, information on whether occupied (non-drivable) or unoccupied (drivable) is indicated.
% This map is shown in the red box located in the upper right of Fig. \ref{fig:system}.
It serves as the input for imitation learning, and it can be used for the \textit{tentacle} and \textit{VVF} algorithms as well.
% It is used as an input for imitation learning and also can be used for the tentacle and velocity vector field (VVF) algorithms.

Driving policies using the occupancy grid map have two advantages.
First, the segmented image can ignore irrelevant information for driving, such as differences in the types of obstacles and pavement in the drivable space.
Therefore, driving policies can achieve similar performance in untrained environments, which can enhance the generality of driving performance.
Second, close and far distance information can be clearly distinguished because the occupancy grid map is a 2D map (i.e., bird's-eye-view).
Thus, depending on the situation, the vehicle can avoid nearby obstacles preferentially or consider distant obstacles in advance.
%%% add

A camera is used to recognize the drivable and non-drivable spaces.
The obstacle detection performance of ultrasonic and 2D-LiDAR sensors depends on the height that they are attached to the vehicle.
With a 3D-LiDAR sensor, the point clouds provide a wide range of height information.
% When using a 3D-LiDAR sensor, a wide range of height information is available through point clouds.
However, this sensor is expensive and requires high computational cost and memory capacity.
In contrast, a camera is compact and inexpensive, and it has lower computational and memory costs.
% The camera sensor is compact and inexpensive compared to the 3D-LiDAR sensor.
% It also requires less computational and memory costs.
In addition, more training data are available for deep learning with vision than with 3D-LiDAR; more data generally help increase the recognition performance.
% Besides, when using deep learning, there are more available training datasets for vision than 3D-LiDAR, which generally shows higher recognition performance.

The upper side of Fig. \ref{fig:system} illustrates the method for obtaining the occupancy grid map from a camera.
% A method to obtain the occupancy grid map using the vision data is illustrated in the upper side of Fig. \ref{fig:system}.
The distortion of the front view camera image is corrected by using intrinsic and extrinsic parameters, but slight distortion remains in the side of the image.
Nevertheless, a trapezoid area of this image has little distortion and is transformed into a bird's-eye-view image through the warp perspective function of the OpenCV library.
The transformed image is segmented into the drivable space and non-drivable space with a deep neural network through semantic segmentation, which refers to the process of linking each pixel to a class label.
The following paragraph describes this network in detail.
The road, crosswalk, and road marks are labeled as the drivable space.
The outside area excluding the drivable space is considered non-drivable space.
The road boundary lines, sidewalks, parking spaces (including parking lines), pedestrians, and vehicles are also labeled as the non-drivable space.
The 200$\times$200 segmented image is divided into 8$^2$ pixels per one grid to obtain a 25$\times$25 grid map.
If all pixel values inside each grid are non-occupied, the grid is regarded as non-occupied.

%%% The perception network in Fig. \ref{fig:system} illustrates the deep neural network structure used to obtain the occupancy grid map.
% The deep neural network structure used for obtaining the occupancy grid map is illustrated in the perception network of Fig. \ref{fig:system}.
The perception network is similar to the segmentation task of MultiNet \cite{teichmann2018multinet}, which is based on the U-Net structure.
It consists of an encoder and decoder based on a convolutional neural network (CNN).
% It consists of the convolutional neural network (CNN) based encoder and decoder.
The encoder is the same as that of the VGG network \cite{simonyan2014very} except for the last layer.
% The encoder is the same as the VGG network \cite{simonyan2014very}, except for the last layer. 
It consists of five pairs of convolutional and max-pooling layers, which is used to extract several abstract features from the input image.
Then, one 1$\times$1 fully-connected layer is connected at the end.
The structure of the decoder follows that of the mainstream fully convolutional network.
The output of the encoder is passed through a 3$\times$3 convolutional layer and up-sampled with three transposed convolution layers.
At this time, each convolutional layer of the encoder is combined with the decoder through the skip connections to extract high-resolution features from the encoded low-resolution features.

\subsection{Imitation Learning for Autonomous Driving in Unstructured Environment}
\label{subsec:imitation_learning}
Imitation learning involves imitating the behavior of an expert for a certain state.
State action pairs of data are collected while an expert is driving.
% State-action pair datasets are collected by an expert driving.
The policy $\pi_{net}$ (i.e., deep neural network) is trained with the data in a process called \textit{behavior cloning}, which is a basic training step of imitation learning \cite{bojarski2016end}.
% A function approximator $\pi_{net}$ (deep neural network) is trained using these datasets.
% This process is called \textit{behavior cloning} which is a basic training step of imitation learning \cite{bojarski2016end}.
To address the limitations of \textit{behavior cloning}, \textit{DAgger} \cite{ross2011reduction} is used to collect additional data by executing the trained \textit{behavior cloning} policy and retraining $\pi_{net}$.
% To address limitations of \textit{behavior cloning}, the \textit{DAgger} \cite{ross2011reduction} algorithm is used.
% \textit{DAgger} collects additional datasets by executing the trained \textit{behavior cloning} policy and retrains $\pi_{net}$.
This process is repeated until the best policy is obtained.
The following subsections describe the composition and collection of the dataset, \textit{behavior cloning}, and the \textit{DAgger} algorithm.
% In addition, the \textit{decision rule} in \textit{DAgger} for effectively collecting the additional data is explained.

    \subsubsection{Dataset}
    \label{subsub:dataset}
    The dataset consists of state and action pairs $D=\{(s_t, a_t)\}_t$, where $t$ is an index of the data.
    The state $s_t$ is the occupancy grid map (25$\times$25 grid $\in$ \{\textbf{\textit{0}} (black): drivable(unoccupied), \textbf{\textit{1}} (white): non-drivable(occupied)\}).
    It is used for the input of the policy $\pi_{net}$.
    
    The action $a_t$ is a command of an expert and the mean value of the output of $\pi_{net}$.
    In this study, the look-ahead point was used as the action $a_{t}$ $\in$ \{$a_{t_x}$, $a_{t_y}$\}, which is the target waypoint for a vehicle to reach.
    % , where $a_{t_x}$ and $a_{t_y}$ are the Cartesian coordinates of the state $s_t$.
    Most autonomous driving studies based on imitation learning use the steering-accel/brake as the action, but the look-ahead point is more useful for executing the proposed \textit{DAgger} algorithm.
    This is explained in detail in Section \ref{subsub:reason_look-ahead_point}.
 
    The output of the policy $\pi_{net}$ for a state is expressed as:
    \begin{equation}
        a_{net, t} = \pi_{net}(s_t),
        \label{eq:state-action}
    \end{equation}
    which consists of $a_{net, t}$ $\in$ \{$\bar{a}_{net, t_x}$, $\bar{a}_{net, t_y}$, $\sigma^2_{a_{net, t_x}}$, $\sigma^2_{a_{net, t_y}}$\}, where are the mean and variance of the look-ahead point.
    The variance of the look-ahead point is calculated through a Gaussian process (GP) to quantify the uncertainty or confidence of $\pi_{net}$ \cite{kendall2017uncertainties}.
    % The variance of the look-ahead point is calculated using Gaussian Process (GP) \cite{rasmussen2003gaussian} to measure an uncertainty or confidence of $\pi_{net}$ quantitatively \cite{kendall2017uncertainties}.

    \begin{figure}[t]
    \begin{center}
        \subfigure[]{\includegraphics[width=0.45\linewidth]{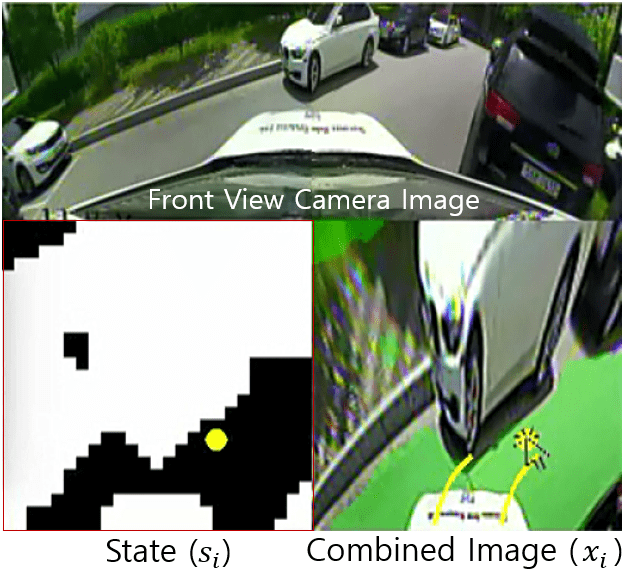}
        \label{fig:BC_a}}
        \subfigure[]{\includegraphics[width=0.45\linewidth]{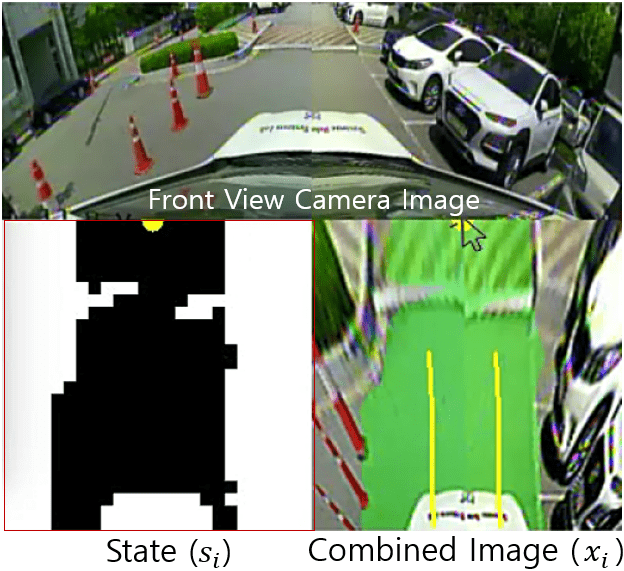}
        \label{fig:BC_b}}
    \end{center}
    \caption{ 
        Dataset collection process of imitation learning (behavior cloning).
        The yellow point is the action $a_{exp, t}$ selected by an expert.
        The expert selects $a_{exp, t}$ in the combined image $x_t$.
        The yellow lines are the future trajectory that the vehicle will drive towards $a_{exp, t}$ along during a certain time.
        The state $s_t$ is the occupancy grid map. 
        The white area of the grid represents obstacles, and the black area represents the drivable space.
    }
    \label{fig:BC}
    \end{figure} 
    
    To collect training data, the expert selects the look-ahead point $a_{exp, t}$ $\in$ \{$a_{exp, t_x}$, $a_{exp, t_y}$\}, and the vehicle is controlled to reach the selected look-ahead point in real-time.
    The steering angle command is calculated with the pure pursuit algorithm \cite{ahn22accurate}.
    The velocity command is proportional to the distance between the look-ahead point and the vehicle.
    As the vehicle is driving, the dataset $D = \{(s_t, a_{exp, t})\}_t$ is stored for every period $t$, and numerous data can be collected easily.
    % While driving, the dataset $D = \{(s_t, a_{exp, t})\}_t$ is stored in every period $t$, and numerous datasets can be collected.
    This process is repeated continuously until the driving is completed.
    
    As illustrated in Fig. \ref{fig:BC}, the expert selects the look-ahead point $a_{exp, t}$ by using a mouse pointer in the combined image $x_t$ instead of the state $s_t$ (i.e., occupancy grid map): 
    \begin{equation}
        a_{exp, t} = \pi_{exp}(x_t),
        \label{eq:state-expert}
    \end{equation}
    where $\pi_{exp}$ indicates the behavior of the expert.
    The combined image $x_t$ is an image of transparently combining the information of the drivable space to the \textit{RGB} image: $x_t$ $\in$ \{\textit{RGB with green}: drivable, \textit{RGB only}: non-drivable\}.
    % The combined image $x_t$ is an image that information of the drivable space is transparently combined in the bird's-eye-view image (\textit{RGB}); $x_t$ $\in$ \{\textit{RGB with green}: drivable space or \textit{RGB only}: non-drivable space\}.
    %%% The look-ahead point is selected in $x_t$ instead of the state $s_t$ (i.e., occupancy grid map) 
    This is because, if $s_t$ is inaccurate (i.e., noisy), the expert may incorrectly select the look-ahead point \cite{kelly2019hg}. 
    % The reason that the expert select the look-ahead point on $x_t$ instead of the state $s_t$ (i.e., occupancy grid map) is that if the perception result of $s_t$ is not accurate (noisy), the expert may incorrectly select the look-ahead point \cite{kelly2019hg}. 
    This situation is shown in Figs. \ref{fig:BC_b}, \ref{fig:BC_problem}, and \ref{fig:noisy}.
    
    The look-ahead point has a geometric relationship with the combined image $x_t$, and the expert $\pi_{exp}$ selects the look-ahead point $a_{exp, t}$ by referring to three criteria (rules):
    % The expert $\pi_{exp}$ selects the look-ahead point $a_{exp, t}$ with three criteria (rules):
    \begin{enumerate}[label=(\roman*)]
        \label{enumerate:criteria}
        \item The look-ahead point must be within the drivable space.
        \item When an obstacle is in front of the vehicle, the expert selects a look-ahead point for which obstacle avoidance is possible.
        % When the obstacle exists in front of the vehicle, the expert selects the look-ahead point where obstacle avoidance is possible. 
         
        To easily check that selected the look-ahead point follows this criterion, the expert can refer to a future trajectory that the vehicle will drive along during a certain time.
        The look-ahead point is selected with the fewest obstacles around the trajectory.
        % In order to easily check that the look-ahead point selected by the expert follows this rule well, the expert refers to a future trajectory that the vehicle will drive during a certain time.
        This trajectory is obtained by using the kinematic model of the vehicle, indicated in Figs. \ref{fig:BC} and \ref{fig:BC_problem}.
        % , where the steering angle and velocity are calculated according to the selected look-ahead point.
        % This trajectory is obtained by using the vehicle kinematic model with steering angle and velocity calculated according to the selected look-ahead point.
        % The look-ahead point is selected where there are the fewest obstacles around the trajectory and the direction of the trajectory.
        \item If there is no obstacle in front of the vehicle, the look-ahead point is selected as far as possible from the vehicle within the drivable space.
        % If the obstacle does not exist in the front of the vehicle, 
    \end{enumerate}
    %%% The look-ahead point can have a geometric relationship for the state through these criteria, so the action pattern for the state can be more clearly found than the steering angle-velocity...
    Based on these rules, the vehicle can avoid obstacles and drive toward the drivable space as fast as possible.
    For example, if an obstacle exists on the front and left side of a vehicle, the look-ahead point is selected to be on the right and near the front side of the vehicle in the drivable space (see Fig. \ref{fig:BC_a}).
    With this point, a large steering angle and low-velocity command are calculated, and the vehicle can safely avoid obstacles.
    Conversely, if there are no obstacles, the look-ahead point is chosen as far as possible from the vehicle in the drivable space (see Fig. \ref{fig:BC_b}).
    With this point, the vehicle can drive at high speed with small steering angle changes.

    \subsubsection{Behavior Cloning}
    \label{subsec:behavior_cloning}
    The collected data can be used to train the policy $\pi_{net}$ in a process similar to that of supervised learning.
    % Using the collected datasets, the process of training the function approximator $\pi_{net}$ is similar to that of supervised learning.
    $\pi_{net}$ is expressed as $\pi_{net}(s_t; \theta)$ parameterized by $\theta$ for the state $s_t$.
    The process of optimizing $\theta$ to minimize the loss function $\mathcal{L}$ is the process of training $\pi_{net}(s_t; \theta)$.
    % The process where $\theta$ is optimized to minimize the loss function $L$ is the process of training $\pi(s_t; \theta)$.
    The loss $\mathcal{L}$ is the difference in the state $s_t$ between the output of $\pi_{net}(s_t; \theta)$ and the action in the dataset $a_{exp, t}$.
    This is expressed as $L(\pi_{net}(s_t; \theta), a_{exp, t})$ , and its detailed expression is given in (\ref{eq:loss}).
    A large number $T$ of datasets $D = \{(s_t, a_{exp, t})\}^N_{t = 1}$ is used to optimize $\theta$.
    This training process is called \textit{behavior cloning} and is expressed as follows:
    \begin{equation} 
        \underset{\theta}{min} \, \sum^T_{t=1} \mathcal{L}(\pi_{net}(s_t;\theta), a_{exp, t}).
        \label{eq:BC}
    \end{equation}
    % This process is called \textit{behavior cloning}.
    The trained policy $\pi_{net}$ can minimize the loss $\mathcal{L}$ with \textit{behavior cloning} policy denoted by $\pi_{BC}$.
    % It becomes a policy of \textit{behavior cloning} denoted by $\pi_{BC}$.
    When a vehicle drives with $\pi_{BC}$ in an environment similar to the trained environment, $\pi_{BC}$ can calculate an action similar to that of the expert.
    
    \begin{figure}[t]
    \begin{center}
        \subfigure[]{\includegraphics[width=0.95\linewidth]{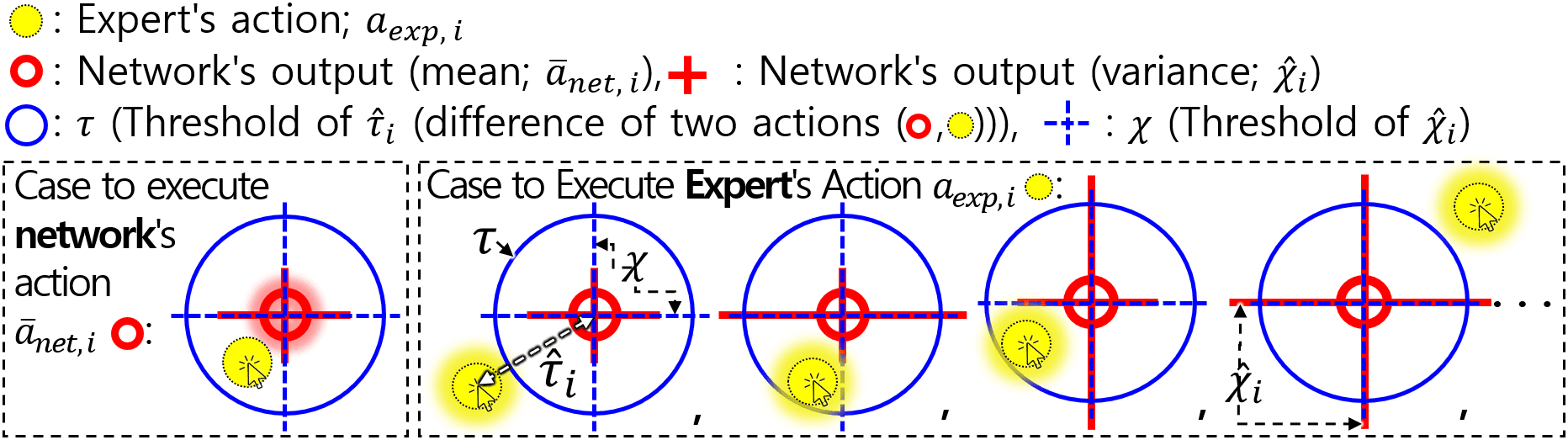}
        \label{fig:bc_DAgger}}
        \subfigure[]{\includegraphics[width=0.95\linewidth]{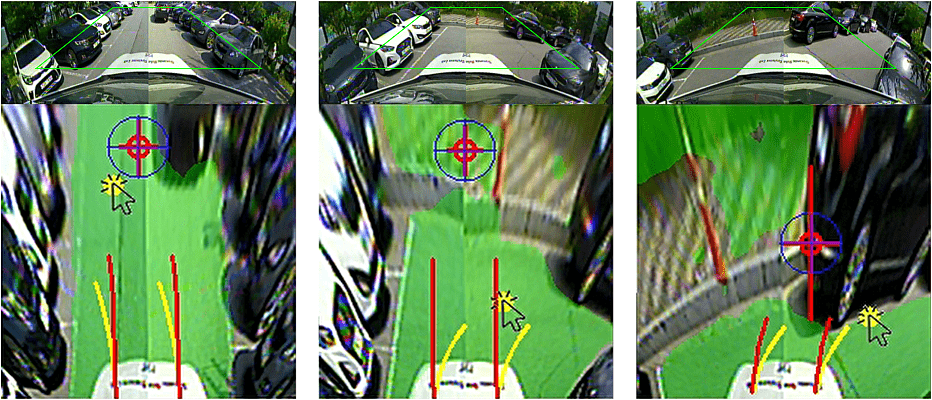}
        \label{fig:BC_problem}}
    \end{center}
    
    % \begin{subfigure}
    %     {\includegraphics[width=0.95\linewidth]{image/4_a notation.png}\label{fig:bc_DAgger}}
    %     % \sidesubfloat \small{(a)}
    % \end{subfigure}
    % \begin{subfigure}
    %     {\includegraphics[width=0.95\linewidth]{image/4_b BC_problem.png}
    %     \label{fig:BC_problem}}
    %     % \sidesubfloat \small{(b)}
    % \end{subfigure}
        \caption{
            Illustration of \textit{DAgger} algorithm.
            (a) Cases where the network's action ($\bar{a}_{net, t}$) or the expert's action ($a_{exp, t}$) is executed.
            (b) unsafe or near-collision situations and the additional dataset collection cases of the \textit{DAgger} algorithm.
            In this example, \textit{DAgger} is in iteration $i$ = 1, and the network $\pi_{net, i=1}$ has the \textit{behavior cloning} policy, $\pi_{BC}$.
            The yellow point is the newly labeled action $a_{exp, t}$ of the expert while $\pi_{BC}$ is being executed.
            The red point in the combined image $x_t$ is the mean of the output by $\pi_{net, i}$: the network's action $\bar{a}_{net, t}$.
            The blue circle is the threshold $\tau$ of $\hat{\tau}_t$ which is the difference between the actions $\bar{a}_{net, t}$ and $a_{exp, t}$.
            The red lines centered at $\bar{a}_{net, t}$ represent the variance of the output of $\pi_{net, i}$: $\hat{\chi}_t$.
            The blue dashed lines centered at $\bar{a}_{net, t}$ represent the threshold of $\hat{\chi}_t$ which is the variance of the output of the network $\pi_{net, i}$: $\chi$.
        }
    \end{figure}
    
    On the other hand, if $\pi_{BC}$ encounters states that are not similar to the dataset $D$ or are noisy, $\pi_{BC}$ may produce unsafe or unsafe actions.
    The noisy state is when the boundary of the drivable space or shadow area is not accurately recognized.
    % The noisy state is when the recognition is not accurate at the boundary of the drivable space or shadow area.
    This result is shown in Fig. \ref{fig:BC_problem} and is the result of executing $\pi_{BC}$ on different days in a place used to collect dataset for \textit{behavior cloning}.
    However, the location or type of obstacle differs from when the dataset was collected for $\pi_{BC}$.
    In this case, the vehicle cannot sufficiently avoid obstacles; this is called the \textit{data mismatch problem}.
    When a vehicle enters a narrow corner, if the non-drivable space (obstacle) is erroneously recognized as the drivable space, the look-ahead point can be placed on the non-drivable space, and the vehicle will collide with the obstacle.
    This occurs when the data for these situations are included in the training dataset $D$ less often than situations of driving in a relatively large drivable space or with no misrecognition problems.
    % The reason is that the datasets for these situations were included in the training dataset $D$ less than those driving in relatively large drivable space or having no misrecognition problems (not noisy).
    Thus, the policy $\pi_{net}$ cannot reflect these situations in $\pi_{BC}$ well; this is called the \textit{data imbalance problem}.
    Moreover, when these problems occur in a driving situation, the error may magnify afterward because $\pi_{BC}$ has not learned recovery behavior; that is called the \textit{compounding error problem}.
    % Moreover, when these problems occur in a certain driving situation, these will also occur after this situation because the $\pi_{BC}$ has not learned a recoverable behavior; \textit{compounding error problem}.
    
    \subsubsection{DAgger Algorithm}
    \label{subsec:DAgger}
    The \textit{DAgger} algorithm can be used in imitation learning to address the problems of \textit{behavior cloning} \cite{ross2011reduction}.
    % To address the problems of \textit{behavior cloning}, one of the imitation learning algorithms, the \textit{DAgger} algorithm is used \cite{ross2011reduction}.
    \textit{DAgger} aggregates an additional dataset $D_i$ with the previously collected dataset $D$ and trains the policy $\pi_{net}$ again. % repeatedly
    This process is repeated until the desired policy is obtained.
    \textit{DAgger} is explained in detail in Algorithm \ref{algo:dagger}.
    
    % DAGGER
    \begin{algorithm}[h]
        \textbf{function} DAgger($\pi_{BC}$, $D_{BC}$)\\
        Initialize $\pi_{net, 1}$ $\leftarrow$ $\pi_{BC}$ \\
        Initialize $D$ $\leftarrow$ $D_{BC}$ \\
        Initialize $i$ $\leftarrow$ 1, $\hat{\eta}_i$ $\leftarrow$ 0.0 \\
        \While{$\hat{\eta}_i$ $\leq$ $\eta$} {
            $D_i$, $\hat{\eta}_i$ $\leftarrow$ Sample unsafe or near-collision datasets using \textbf{\textit{Data-sampling Function}}($\pi_{net, i}$) \\
            Aggregate datasets $D$ $\leftarrow$ $D$ $\bigcup$ $D_i$ \\
            Train policy $\pi_{net, i+1}$ on $D$ using Eq. \ref{eq:BC} \\
            $i$ += 1 \\
        }
        \textbf{return} $\pi_{net, i}$ 
    \caption{Pseudo-code of \textit{DAgger} Algorithm}
    \label{algo:dagger}
    \end{algorithm}
    First, \textit{DAgger} initializes the policy $\pi_{net, i=1}$ and dataset $D$ as those obtained by \textit{behavior cloning}.
    The \textit{DAgger} iteration $i$ and $\hat{\eta}_i$ representing the performance of the trained policy $\pi_{net, i}$ are initialized.
    When the iteration is started ($i$ = 1, line 6), the additional dataset $D_i$ is collected by the \textit{data-sampling function} as described in the next subsection.
    % When starting the iteration ($i$ = 1), the additional dataset $D_i$ is collected by the \textit{decision rule} as described in the next subsection \ref{subsec:Decision_Rule}.
    The \textit{data-sampling function} checks whether an unsafe or near-collision situations occurs.
    %%% With the \textit{decision rule}, the action of $\pi_{net, i}$ and the expert action are calculated simultaneously. 
    % The vehicle is controlled by the action of $\pi_{net, i}$.
    When it occurs, the expert action is used to control the vehicle, and the state and action are collected in the additional dataset $D_i$ (see Fig. \ref{fig:BC_problem}).
    Otherwise, the action of $\pi_{net, i}$ is used to control, and the additional dataset is not gathered.
    
    After the driving takes place, the collected additional dataset $D_i$ is aggregated to the existing dataset $D$ (line 7).
    The aggregated dataset $D$ is used to retrain the policy $\pi_{net}$ with (\ref{eq:BC}) (line 8).
    After training, a policy $\pi_{net, i+1}$ that can cause fewer unsafe or near-collision situations than $\pi_{net, i}$ can be obtained.
    As more data of these problem situations are aggregated, $\pi_{net, i}$ becomes more capable of dealing with these situations, which is proven in \cite{ross2011reduction}.
    \textit{DAgger} repeats this process until these problem situations rarely happen (line 5).
    This can be judged by $\hat{\eta}_i$ (line 19 of Algorithm \ref{algo:Decision_rule}) which is the ratio of executed network actions among the total executed actions.
    % which is the proportion of executed network actions.
    If $\hat{\eta}_i$ is greater than the threshold $\eta$, the iterations of \textit{DAgger} are terminated.
    Finally, a policy $\pi_{net, i}$ that does not cause unsafe or near-collision situations is obtained (line 11).
    
    \subsubsection{\textit{Data-sampling Function} in \textit{DAgger}}
    \label{subsec:Decision_Rule}
    To collect the additional dataset $D_i$ and judge the performance of the trained policy, the \textit{data-sampling function} determines whether to use the trained policy or expert behavior depending on the driving situation.
    % The \textit{decision rule} used in this study is based on \textit{EnsembleDAgger} \cite{menda2019ensembledagger}.
    To imitate the additional dataset more precisely and quickly, the error defined by the EnsembleDAgger algorithm \cite{menda2019ensembledagger} is reflected in the training process.
    % Additionally, to more precisely imitate dataset, we propose that an error defined in \textit{EnsembleDAgger} is reflected in the training process.
    The \textit{data-sampling function} is explained in detail in Algorithm \ref{algo:Decision_rule}.
    
    \begin{algorithm}[t]
        \textbf{function} \textit{\textbf{Data-sampling Function}}($\pi_{net, i}$)\\
        Initialize $D_{i}$ $\leftarrow$ $\emptyset$ \\
        Initialize $n_{tot}$ $\leftarrow$ 0, $n_{net}$ $\leftarrow$ 0 \\
        \For{$t$ = 0 \textbf{to} End of Execution} {
            $\bar{a}_{net, t}$, $\sigma^2_{a_{net, t}}$ $\leftarrow$
            $\pi_{net, i}(s_t)$ \\
            $a_{exp, t}$ $\leftarrow$ $\pi_{exp}(x_t)$ \\
            $\hat{\tau}_t$ $\leftarrow$ $\|\bar{a}_{net, t} - a_{exp, t}\|_2$ \\
            $\hat{\chi}_{t_{j\in{x, y}}}$ $\leftarrow$ $\sigma^2_{a_{net, t}}$ \\
            \If{$\hat{\tau}_t$ < $\tau$ \textbf{or} $\hat{\chi}_{t_x}$ < $\chi$ \textbf{or} $\hat{\chi}_{t_y}$ < $\chi$} {
                Control the vehicle with $\bar{a}_{net, t}$ \\
                $n_{net}$ += 1 \\
            }
            \Else{
                Control the vehicle with $a_{exp, t}$\\
                $D_i$.append(\{$s_{t}$, $a_{exp, t}$, $\hat{\tau}_t$\}) \\
            } 
            $n_{tot}$ += 1 \\
        }
        $\hat{\eta}_i$ $\leftarrow$ $\dfrac{n_{net}}{n_{tot}}$ \\
        \textbf{return} $D_{i}$, $\hat{\eta}_i$
    \caption{Pseudo-code of \textit{Data-sampling Function} in \textit{DAgger} Algorithm}
    \label{algo:Decision_rule}
    \end{algorithm}
    
    The \textit{data-sampling function} quantifies the similarity and confidence for the output of the trained policy $\pi_{net, i}$. 
    This can be used to determine quantitatively whether the driving situation of $\pi_{net, i}$ is unsafe or a near-collision.
    % Using them, it is quantitatively determined whether the situation driven by the $\pi_{net, i}$ is the unsafe or near-collision.
    The outputs of $\pi_{net, i}$ and the expert behavior $\pi_{exp}$ are obtained simultaneously
        % \footnote{ If the action is the steering angle and velocity, the expert action cannot be obtained at the same time as when the vehicle is being driven with the trained network.
        % On the other hand, because the look-ahead point was used as the action in this study, the expert can select it with a mouse pointer on the combined image (RGB + segmented) regardless of the network action.}
    (lines 5 and 6) and compared before either is used to control the vehicle (lines 7-9).
    
    The discrepancy (error) between the actions of $\pi_{net, i}$ and $\pi_{exp}$ is calculated to check the similarity between the two actions (line 7).
    % The discrepancy of the actions between $\pi_{net, i}$ and $\pi_{exp}$ (error) is calculated to check the similarity of the two actions; $\hat{\tau}_t$ $\in$ [0, 1] (line 7).
    To quantify the confidence of $\pi_{net, i}$, the variance of $\pi_{net, i}$ is obtained: $\hat{\chi}_t$ (line 8). 
    % To measure the confidence of $\pi_{net, i}$ quantitatively, the variance of $\pi_{net, i}$ is obtained; $\hat{\chi}_t$ (line 8). 
    As shown in Fig. \ref{fig:bc_DAgger}, by checking whether $\hat{\tau}_t$ or $\hat{\chi}_t$ is larger than threshold values $\tau$ or $\chi$, unsafe or near-collision situation can be identified.
    
    As shown in Fig. \ref{fig:BC_problem}, in all three situations, $\hat{\tau}_t$ is greater than $\tau$ (blue circle). 
    In the rightmost case, $\hat{\chi}_t$ (red lines) is also greater than $\chi$ (blue lines).
    In these cases, if the vehicle follows the action of the network (red circle), the distance between the vehicle and obstacle decreases, and the possibility of collision increases.
    To avoid unsafe situations in these cases, the action of the expert behavior (yellow circle) is used to control the vehicle (line 14).
    At the same time, only the state $s_t$ of this situation and the expert action $a_{exp, t}$ are collected to the additional dataset $D_i$ (line 15).
    This is to train the network intensively to overcome unsafe and near-collision situations.
    % This is to train the network to overcome the unsafe or near-collision situations intensively.
    
    By using the criteria for $\hat{\tau}_t$ and $\hat{\chi}_t$ (line 9), the states with unsafe or near-collision situations can be collected as much as possible within a range where the vehicle does not collide with obstacles.
    If the expert judges these situations heuristically without using these criteria, these problem states cannot be sufficiently collected.
    This is because experts prefer to avoid these situations immediately, so they are difficult to experience them.
    In the next iteration $i+1$, these situations can be handled better with a larger dataset containing these problem situations, in contrast to when the criteria are not used.
    
    The discrepancy between actions $\hat{\tau}_t$ is then added to the additional dataset $D_i$ to imitate the training dataset more precisely (line 15).
    $\hat{\tau}_t$ is reflected in the training process as a weight (gain) for the loss function $\mathcal{L}$, as expressed in (\ref{eq:mse_gain}).
    This updates the parameters of the network as much as the network
    generates the error ($\hat{\tau}_t$),
    % incorrectly computes the action, 
    which can reduce the possibility of the same mistake being repeated.
    Therefore, at the next iteration $i+1$, \textit{DAgger} can reduce $\hat{\tau}_t$ more with the weight than without it.
    Thus, the final policy $\pi_{net, i}$ can be obtained with fewer \textit{DAgger} iterations.
    
    %%%%%%%%%%%%%%%%%%%%% add
    \subsubsection{Reasons to Use Look-ahead Point As Action}
    \label{subsub:reason_look-ahead_point} 
    If the action is the steering-accel/brake, the expert suffers two problems in executing the \textit{DAgger} algorithm, and these can be addressed by using the look-ahead point.
    
    First, as shown in lines 5 and 6 of Algorithm \ref{algo:Decision_rule}, the network action and expert behavior should be obtained simultaneously.
    However, if the steering-accel/brake is used as the action, an expert action cannot be obtained at the same time when the vehicle is being controlled by a network action.
    On the other hand, because the proposed method uses the look-ahead point as the action, the expert can select the look-ahead point with only a mouse pointer on the combined image $x_t$ regardless of the network action. %(RGB + segmented)
    
    Second, even if the action is set as the steering-accel/brake and the expert action can be obtained simultaneously with the network action, the expert cannot clearly find a steering-accel/brake value that the vehicle can drive as safe and fast as possible when performing \textit{DAgger}.
    This is because, when the vehicle is controlled by the network and expert intervention is needed, the expert cannot calculate an action value considering the current network action used for vehicle control.
    When humans drive, they do not directly calculate an absolute steering-accel/brake value, but calculate how much more or less rotate the steering angle and press the accel/brake pedals from the current steering-accel/brake (i.e., amount of change).

    In this study, the expert selects the look-ahead point that the vehicle will reach on the combined image $x_t$ by referring to the three criteria mentioned in the previous subsection \ref{enumerate:criteria}).
    These criteria specify where the look-ahead point is chosen for $x_t$ by its geometric relationship.
    Thus, the expert can clearly find one look-ahead point that the vehicle can drive as safe and fast as possible without the current steering-accel/brake feedback of the vehicle controlled by the network.
    This enables a state-action relationship (pattern) to be clearly identified, so a neural network can learn the driving pattern more clearly.
   
    \subsubsection{Driving Policy Network}
    \label{subsub:driving_policy_network}
    The deep neural network is used as the policy $\pi(s_t; \theta)$, which is illustrated in the driving policy network of Fig. \ref{fig:system}.
    It consists of the encoder with the CNN and fully connected layers.
    The encoder is composed of two pairs of convolutional and max-pooling layers, and the flattened layer nodes are connected.
    Then, the two fully connected layers with 1000 and four nodes are linked at the end.
    The last layer with four nodes has the mean and variance for $x$ and $y$ of the look-ahead point.

    The loss function $\mathcal{L}(\pi_{net}(s_t; \theta), a_{exp, t})$ in (\ref{eq:BC}) is the multivariate Gaussian log-likelihood loss function (see (\ref{eq:loss})).
    This allows the network to infer the mean and variance of the Gaussian distribution for the output \cite{kendall2017uncertainties}.
    \begin{equation} 
        \mathcal{L} = \dfrac{1}{n} \sum_{j\in{x,y}} \frac{1}{2} {r_j}^T K_{r_j}^{-1} r_j + \frac{1}{2} \log |K_{r_j}|,
        % - \dfrac{n}{2}\log(2\pi),
        \label{eq:loss}
    \end{equation}
    where the first term penalizes wrong predictions.
    The second term predicts model complexity and penalizes it.
    % , and the third constant term is for normalization.
    $n$ is the dimension of the action, and the look-ahead point has two dimensions: the $x$ and $y$ axes.
    The variable $diff_j$ is the difference between the expert action $a_{exp, t}$ in the dataset and the predicted action by the network $\pi(s_t;\theta)$ which is the network output being trained for $s_t$:
    \begin{equation} 
        diff_{j \in \{x, y \} } = |a_{exp, t_j} - \pi(s_t; \theta)_j|.
    \label{eq:mse}
    \end{equation}
    In addition, the weight (gain) $\hat{\tau}_t$ is multiplied by $diff_j$ for more effective learning \cite{byrd2019effect}, which is described in the previous subsection (\ref{subsec:Decision_Rule})):
    \begin{equation} 
        r_j = (1 + \alpha \hat{\tau}_t) \, diff_j,
    \label{eq:mse_gain}
    \end{equation}
    where $\alpha$ is the gain of $\hat{\tau}_t$.
    During the training process of \textit{behavior cloning}, because there is no $\hat{\tau}_t$ in the dataset, $\alpha \hat{\tau}_t$ is not reflected in (\ref{eq:mse_gain}).
    The effectiveness of applying the weight $\hat{\tau}_t$ in $diff_j$ is analyzed in the fourth row of Table \ref{table:data}.
    $K_{r_j}$ in (\ref{eq:loss}) is $\sigma^2_{a_{net, t_j}}$, which is used to estimate the variance of the output of the network $\pi(s_t; \theta)$.
    % a $s$$\times$$s$ dimensional covariance matrix which includes the model parameters; 
    
\section{Experimental Setup}
\label{sec:experimental_setup}
% This section describes the autonomous vehicle system, and the detailed (training) setup of the perception and driving policy components presented in Fig. \ref{fig:system}.

\subsection{Vehicle and Hardware}
\begin{figure}[h]
\centering
	\includegraphics[width = 0.73\linewidth]{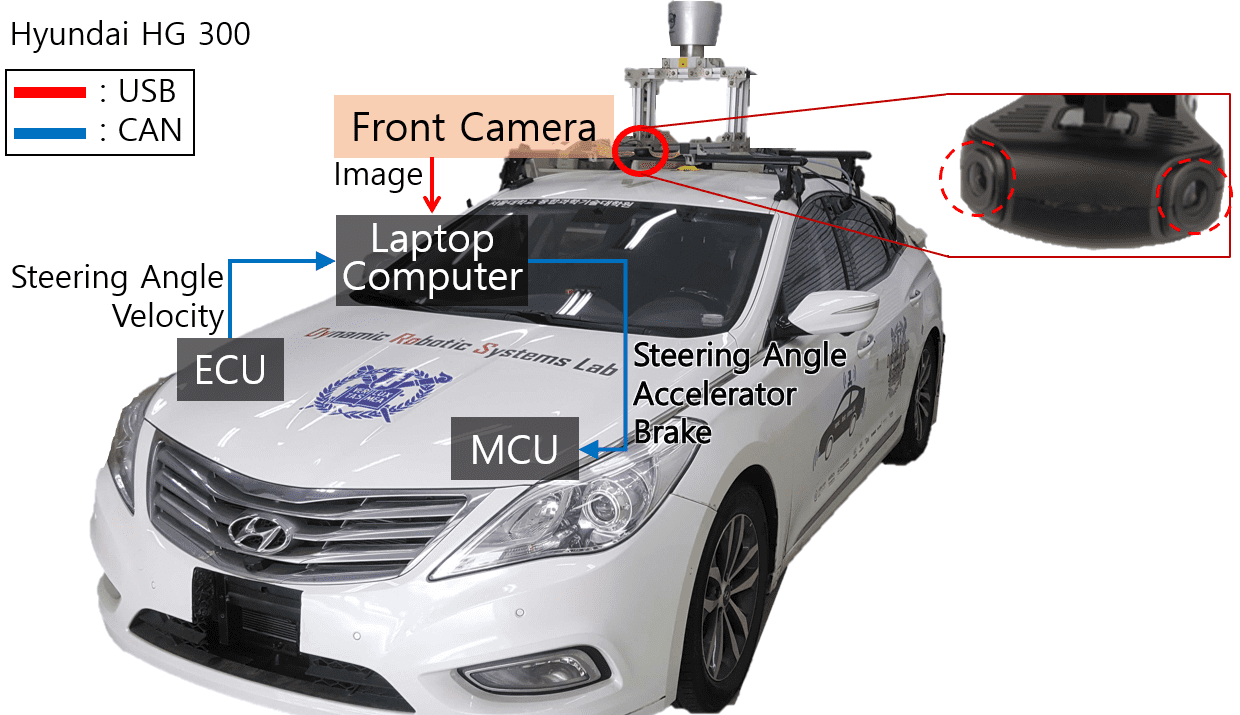}
    \caption{
        Hardware System.
        % Autonomous vehicle hardware system architecture
    }
\label{fig:vehicle}
\end{figure} 
The vehicle used in the experiments was a Hyundai HG 300, as shown in Fig. \ref{fig:vehicle}.
The operating system of the laptop computer was Ubuntu 16.04, and the robot operating system (ROS) was used as a meta-OS platform.
The GPU was Nvidia GTX 1080-ti (8 GB), and the CPU was 3.9 GHz Intel i9-8950HK.
The steering wheel, accelerator, and brake were controlled by a micro controller unit using a proportional-integral-derivative (PID) controller.
 
% \subsubsection{Camera}
A front camera was attached 1.55 m above the ground and 0.25 m forward from the vehicle center. 
It was rotated about 20$^\circ$ in the pitch direction to minimize the shaded area of the bird's-eye-view image.
This camera comprised two lenses to capture a wide view of the environment (see Fig. \ref{fig:vehicle}). 
The field of view (FoV) of each lens was 120$^\circ$, and the distortion of images was corrected.
% Two images obtained were concatenated and shown in Fig. \ref{fig:system}.

\begin{figure*}[t]
    \begin{center}
        \subfigure[]{\includegraphics[width=0.39\linewidth]{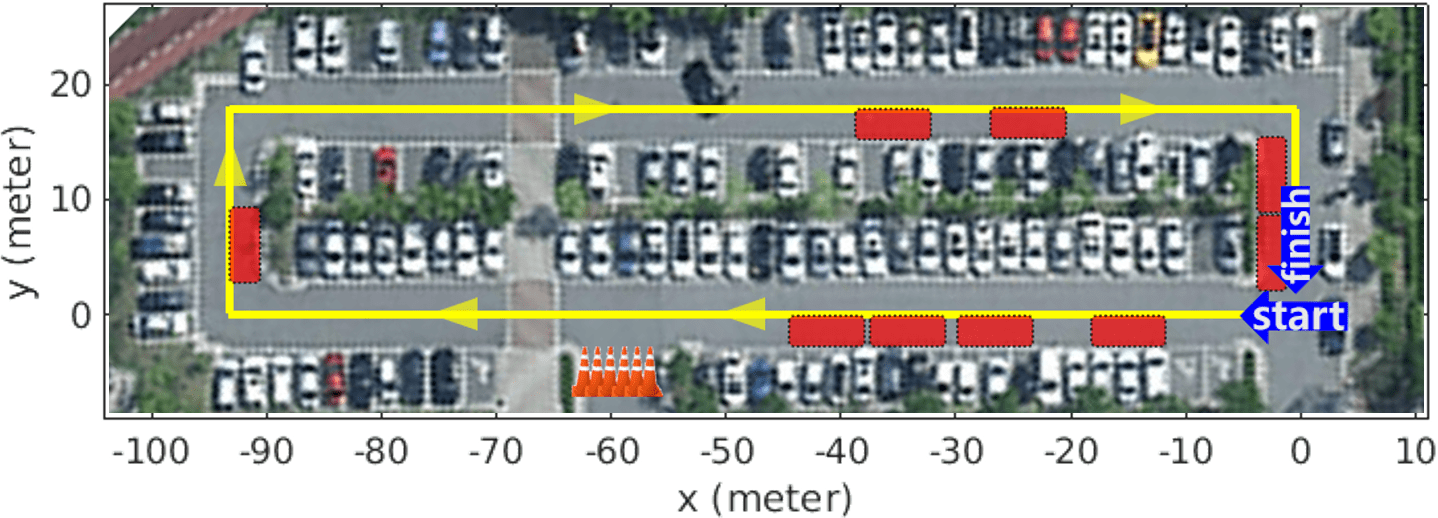}\label{fig:map1}}
        \subfigure[]{\includegraphics[width=0.26\linewidth]{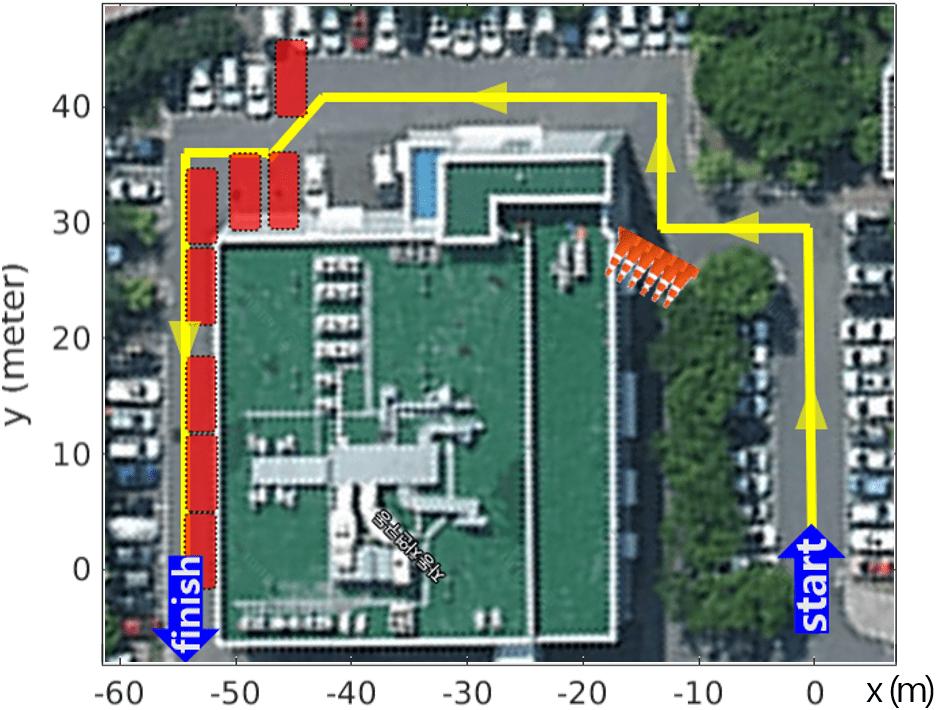}\label{fig:map2}}
        \subfigure[]{\includegraphics[width=0.34\linewidth]{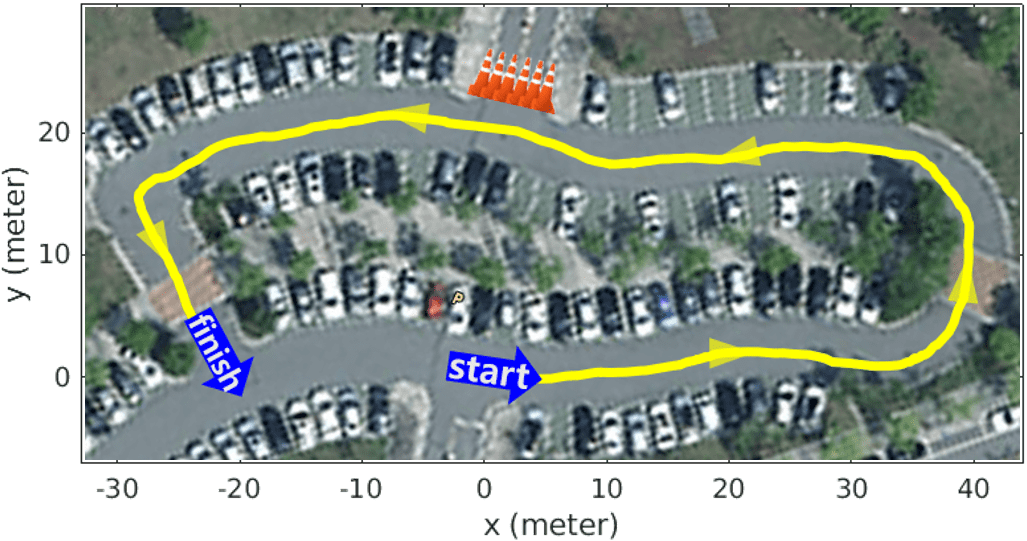}\label{fig:map3}}
    \end{center}
\caption{
Parking lots used in the experiment.
At intersections, traffic cones are used to guide vehicles to drive in one direction.
The yellow line is the center of the drivable space.
The red boxes represent obstacle vehicles that were present in the fifth experiment.
(a) Yellow line is about 230 m long; this parking lot was used to collect the training dataset for imitation learning. 
(b) Yellow line is 139 m long.
(c) Yellow line is 149 m long.
}
\label{fig:map}
\end{figure*}
% \subsubsection{Control}
% A vehicle followed the look-ahead point obtained from the driving policy network.
The pure pursuit algorithm \cite{ahn22accurate} was used to calculate the steering angle command ($\delta$) to reach the look-ahead point:
$\delta$ = $\tan^{-1}\Big(\frac{2L\sin\theta_l}{L_f}\Big)$,
where $L$ is the wheelbase, and $L_f$ is the distance between the positions of the vehicle and look-ahead point.
$\theta_l$ is the look-ahead heading, which is the difference between the heading of the vehicle and heading of the vector from the vehicle to the look-ahead point.
The range of $\delta$ was -540$^\circ$ to 540$^\circ$.

The velocity command $v$ (m/s) to reach the look-ahead point was proportional to $a_y$ which is the longitudinal distance between this point and the vehicle \cite{ahn22accurate}.
Thus, $v$ = $\frac{a_y}{2.24}$, where the final $v$ was set to half of $a_y$ for safety reasons.
The range of $v$ was 0.5 - 2.2 (desired velocity) m/s.
The accelerator and brake commands for controlling the velocity were calculated with the PI controller.

\subsection{Perception Network Training}
Softmax cross-entropy was used as the loss function to train the perception network.
% The loss function to train the perception network was the softmax cross-entropy.
The drivable and non-drivable probability values were inferred for each pixel, and the average loss of each pixel was calculated.
% It inferred the drivable and non-drivable probability values for each pixel, and the average loss of each pixel was calculated.
The Otsu algorithm was used to determine the threshold value about the drivable probability. % \cite{otsu1979threshold}
Before training, weights were assigned to initialize the network for efficient training.
The encoder was initialized with weights trained on ImageNet data.
The transposed convolution layers in the decoder were initialized using the scheme in \cite{teichmann2018multinet} to segment two classes.
% to perform bi-linear up-sampling \cite{teichmann2018multinet}.
The Adam optimizer with a learning rate of 10$^{-5}$ was used to train the network.
A weight decay of 5$\times$10$^{-4}$ was applied to the network.
Epochs were 10k, and the batch size was set to 128.

The training dataset (i.e., RGB-segmented images) was collected for three parking lots as shown in Fig. \ref{fig:map}.
As the vehicle was driven, one image per second was collected for 989 RGB images in total.
The pixel annotation tool \cite{Breheret:2017} was used to segment the collected images into a drivable class and non-drivable class.
The RGB and segmented images were transformed into the bird's-eye-view image and used to train the perception network.
Eighty percent of the dataset was used for training, and the rest was used for validation. 

\subsection{Driving Policy Network Training}
The loss function of the driving policy network is explained in Section \ref{subsub:driving_policy_network}).
Parameter $\alpha$ in (\ref{eq:mse_gain}) was set to 
1.5.
% The average loss for mean and variance of each $x$ and $y$ of the look-ahead point was calculated.
The Adam optimizer with a learning rate of 10$^{-5}$ was used to train the network.
The network was not initialized with pretraining weights. 
Epochs were 100k, and the batch size was set to 512.

The training dataset $D$ was collected for only the parking lot shown in Fig. \ref{fig:map1}.
The vehicle was driven from the start point to the finish point.
To collect more data, the vehicle was turned around from the finish point and driven to the start point (totaling 460 m).
% Additionally, to collect more data, it turned around from the finish point and drove to the start point (totaling 460 m).
The dataset was collected according to the method explained in Section \ref{subsub:dataset}), and the process was recorded as a video \footnote{\url{https://youtu.be/KOXFTEYL-xs}}.
% \footnote{\url{https://youtu.be/su5o_sXRi04}}.
% The method to collect the dataset is explained in \ref{subsub:dataset}.

The threshold $\tau$ of $\hat{\tau}_t$ (discrepancy of two actions) in Algorithm \ref{algo:Decision_rule} was set to 0.07.
$\tau$ was set to the maximum value at which the vehicle cannot collide with an obstacle by the action of the network when performing \textit{DAgger}.
This is because the higher this value, the more unsafe or near-collision data can be obtained.
The threshold $\chi$ of $\hat{\chi}_{t_{j\in{x, y}}}$ (variance of the action) was set to 0.1.
The threshold $\eta$ of $\hat{\eta}_i$ (performance of the trained policy, line 19 in Algorithm \ref{algo:Decision_rule}) was set to 0.9.

\begin{table}[h]
% \caption { }
\caption {\textit{DAgger} Results}
\begin{adjustbox}{width=1.0\linewidth}
\small
\label{table:data}
    % \begin{center}
\begin{tabular}{c|c|c|c|c|c|c}
\Xhline{3\arrayrulewidth}
& \textit{BC} & $i$ = 1 & 2 & 3 & 4 & 5 \\ \hline \hline
Dataset (ea) & 6425 & 3258 & 2263 & 2088 & 1505 & 731 \\ \hline
% 257 / 234 / 236 / 258 / 238 / 247
Network/Total ($\hat{\eta}_i$) & - & 0.44 & 0.64 & 0.65 & 0.74 & 0.90 \\ \hline
% rev: 54.5 36.1 35.7 26.9 
Effect of (\ref{eq:mse_gain}) & - & 0.62 & 0.65 & 0.85 & 0.89 & 0.91 \\ \Xhline{3\arrayrulewidth}
\end{tabular}
\end{adjustbox}
% \end{center}
{\\ \footnotesize *\textit{Note}.
\textit{BC} represents for "\textit{behavior cloning}". \\
The "Network/Total" (third) row represents the percentage of network actions among the total actions used in each \textit{DAgger} iteration $i$.
This is identical to 100$\times \hat{\eta}_i$, where $\hat{\eta}_i$ is expressed in line 19 of Algorithm \ref{algo:Decision_rule}. \\
The "Effect of (\ref{eq:mse_gain})" (fourth) row compares the results of accumulating the error ($\hat{\tau}_t$) per second for the state in the dataset between two trained policies that do and do not reflect the weight $\alpha \hat{\tau}_t$ in the loss function: $\dfrac{\Sigma \hat{\tau}_t, which \, reflects \, \hat{\tau}_t \, in \,  (\ref{eq:mse_gain})}{\Sigma \hat{\tau}_t, which \, does \, not \, reflects \, \hat{\tau}_t \, in \, (\ref{eq:mse_gain})}$.
In this row, a lower value indicates a greater difference between the two policies and shows how much the error can be reduced by reflecting $\hat{\tau}_t$ in (\ref{eq:mse_gain}).
% It can be seen that the lower this value is, the greater the difference is between the two policies (with/without reflecting $\hat{\tau}_t$ in (3)).
% This shows how much the error can be reduced by reflecting $\hat{\tau}_t$ in (\ref{eq:mse_gain}).
}
\end{table}

The final policy was obtained through 5 \textit{DAgger} iterations ($i$ = 5).
Increasing the number of \textit{DAgger} iterations can improve performance, but may not not significantly.
% When $\hat{\eta}_i$ exceeded 90 \%, the trained network action was almost similar to the expert action and did not occur any problem.
In the case of our experiment, the difference between $\hat{\eta}_{i=5}$ and $\hat{\eta}_{i=6}$ was only 0.02.
% The total driving time to collect the dataset was 24 min 30 s.
% The total driving time for collecting the dataset was 24 minutes 30 seconds.
As the vehicle was being driven, data were collected at intervals of 0.05 s.
% While driving, one dataset was collected per 0.05 second.
Table \ref{table:data} presents the number of collected data and percentage of executed network actions and the effect of applying $\hat{\tau}_t$ in (\ref{eq:mse_gain}).

\subsection{General Model-based Motion-Planning Algorithms Used for Comparison}
\label{subsub:other_method}
The \textit{tentacle} \cite{mouhagir2019evidential} and \textit{VVF} \cite{wang2015driving} algorithms were used to compare with the proposed method.
These are representative and general algorithms of the candidate path selection and artificial field generation methods, respectively.
The optimization, graph search, and incremental tree search methods could not be considered because they require global information and real-time calculation is difficult.
% The optimization, graph search, and incremental tree search methods mentioned in Section \ref{sec:intro} cannot be considered because these methods require global information, and also real-time calculation is difficult.

Among the candidate path selection algorithms, only \textit{tentacle}'s the curvature of the candidate path gradually increases to account for the constraint that the steering angle cannot be changed by a large amount instantaneously.
Thus, a vehicle can drive more smoothly and safely on a narrow or variable-curvature road with this algorithm than with the dynamic window approach (DWA) \cite{missura2019predictive} and curvature velocity method (CVM) \cite{lopez2019new}.
% So, the vehicle can drive more smoothly and safely on narrow or varied curvature roads than the dynamic window approach (DWA) \cite{missura2019predictive} and curvature velocity method (CVM) \cite{lopez2019new} algorithms.
With regard to artificial field generation, the \textit{VVF} algorithm is similar to the APF algorithm \cite{ge2000new} and can be used without the goal point because it also considers the desired velocity.
% In the artificial field generation method, \textit{VVF} is similar to the artificial potential field (APF) algorithm \cite{ge2000new}, and it can be used without the goal point because it further considers the desired velocity.

The occupancy grid map was used as the input for the \textit{tentacle} and \textit{VVF} algorithms.
The steering angle was calculated with each algorithm, and the velocity was set as inversely proportional to the calculated steering angle.
For example, if the steering angle was zero, the velocity was set to the desired velocity (2.2 m/s); when the steering angle was maximum (540$^\circ$), the velocity was set to the lowest velocity (0.5 m/s).

\subsubsection{\textit{Tentacle} Algorithm \cite{mouhagir2019evidential}}
This algorithm has 16 candidate path sets depending on the velocity, and each candidate path set has 81 candidate paths.
The cost for each candidate path is calculated with the objective function, and the candidate path with the smallest value is selected.
The objective function has four terms: clearance, flatness, trajectory, and forwarding.
The clearance term prefers to choose a candidate path with the fewest obstacles around it.
The flatness term is similar to the clearance term and additionally considers the probability of a grid being occupied (non-drivable) for each cell in the occupancy grid map for smooth driving.
The trajectory term selects a candidate path that can head to the global path.
The forwarding term was further designed in this study to drive forward preferentially.
This term selects the candidate path with the least curvature.
% A candidate path with the least curvature is selected by this term.

In experiments, a set of candidate paths for 2.2 m/s was used.
The application ratio of the clearance, flatness, trajectory, and forwarding terms was; 1:0:0:0.3.
The flatness term was not used because the occupancy grid map in this study did not have the occupied probability.
In addition, the trajectory term could not be used because of the absence of global information.
When the forwarding term was set to greater than 0.3, the oscillation problem was reduced, but the risk of collision was increased for large curvature changes. 
The clearance term included a detection range parameter to calculate the proportion of obstacles around the candidate path.
% The clearance term has the detection range that calculates the proportion of obstacles around the candidate path.
This range was set to 0.35 m, which is the width of the vehicle (0.2 m) plus the safety distance (0.15 m).
When this was increased further, the vehicle could avoid obstacles more safely, but more oscillation occurred in narrow drivable space.

\subsubsection{\textit{VVF} Algorithm \cite{wang2015driving}}
Like the APF algorithm, the VVF algorithm has a repulsive field for obstacles and an attractive field for the goal point.
Additionally, to follow the desired velocity and direction, the velocity field is reflected to the APF field.
To drive along the combined field, the look-ahead point is searched by descending along the gradient of the field's direction from the front of the vehicle.

In experiments, the repulsive, attractive, and velocity fields were set to a ratio of 1:0:0.5.
The attractive field could not be used because global information (global path, localization data) was not used in this study.
The direction of the velocity field was set so that the vehicle could drive forward.
When the fields were combined, only the repulsive field was applied around obstacles with a range of 2.3 m.
% When combining the fields, only the repulsive field was applied around the obstacle, and this range was 2.3 m.
If the range was set greater than 2.3 m, the vehicle could avoid obstacles more safely, but more oscillations occurred when it passed through narrow drivable space.
% If this range was larger than 2.3 m, the vehicle was able to avoid obstacles more safely, but more oscillation occurred when passing through the narrow drivable space.

\section{Experimental Results}
\label{sec:experimental_result}
The experimental results for the perception network and driving policy are presented here to demonstrate the effectiveness of the proposed method.
% To show the performance of the proposed method, this section presents the experimental results of the perception network and the driving policy.
The experiments were conducted at three parking lots without intersections, as shown in Fig. \ref{fig:map}.
There were no lanes in the drivable space, where the width and curvature changed rapidly.
In addition, several unknown static obstacles were present.

In the perception network test, the accuracy and speed of the perception network were measured.
In the driving policy test, the proposed method was compared with the \textit{tentacle} and \textit{VVF} algorithms.
% In the driving policy test, the \textit{Tentacle} and \textit{velocity vector field (VVF)} algorithms and the proposed method were compared.
The driving results were recorded as videos\footnote{\url{https://youtu.be/OQls9fDgiaA}} and quantitatively evaluated according to a designed evaluation metric.
% Driving results were recorded as videos\footnote{\url{https://youtu.be/R_eqPS9co5I}} and evaluated quantitatively by defining an evaluation metric.
The limitations of the \textit{tentacle} and \textit{VVF} algorithms were analyzed for each situation.
In addition, a stability and time-delay of our method were analysed.

\subsection{Perception Network Test Results}
The performance of the perception network was tested with the validation dataset that were not used to train the network. 
% The performance of the perception network was tested by validation dataset images that were not used to train the network. 
The pixel accuracy was used as the evaluation metric:
% It was evaluated using the pixel accuracy metric;
\begin{equation}
    Pixel \,\, Accuracy = \frac{correctly \,\, classified \,\, pixels}{total \,\, number \,\, of \,\, pixels} \, (\%),
    \label{eq:acc}
\end{equation}
where the numerator is the number of pixels correctly predicted by the network.
Table \ref{table:perception} presents the \textit{pixel accuracy} results for the parking lots in Fig. \ref{fig:map}. 

\begin{table}[h]
\centering
\caption {Perception Network Results}
\begin{adjustbox}{width=1.0\linewidth}
\small
\label{table:perception}
\begin{tabular}{l|c|c|c}
\Xhline{3\arrayrulewidth}
\multirow{2}{*}{} & \multicolumn{3}{c}{Parking Lots} \\
 & \multicolumn{1}{c|}{\begin{tabular}[c]{@{}l@{}}Fig. \ref{fig:map1}\end{tabular}} & \multicolumn{1}{c|}{\begin{tabular}[c]{@{}l@{}}Fig. \ref{fig:map2}\end{tabular}} & \multicolumn{1}{c}{\begin{tabular}[c]{@{}l@{}}Fig. \ref{fig:map3}\end{tabular}} \\ \hline \hline 
\begin{tabular}[c]{@{}l@{}}Pixel Accuracy (\ref{eq:acc}) [\%]\end{tabular} & 98.14 & 97.75 & 97.85 \\ \Xhline{3\arrayrulewidth}
\end{tabular}
\end{adjustbox}
\end{table}

The drivable space is represented as the green areas in Figs. \ref{fig:tentacle}-\ref{fig:noisy}.
On average, the execution speed of the perception network was 27.9 fps.

\subsection{Quantitative Analysis of Driving Policy} 
\subsubsection{Collision Rate} 
The \textit{collision rate} was used as an evaluation metric to quantify the performance of each driving policy algorithm.
% A \textit{collision rate} metric is defined to quantitatively analyze the performance of each algorithm.
This metric indicates the number of collisions per 100 m as the vehicle was driven in each parking lot:
\begin{equation}
    \textit{collision rate} = 100 \, \frac{cnt_{col}}{len_{path}},
    \label{eq:collision_rate}
\end{equation}
where $cnt_{col}$ represents the number of times a near-collision situation occurred.
When the vehicle headed toward an obstacle and the distance was 0.5 m or less, the vehicle was stopped, and $cnt_{col}$ was incremented.
% a person stopped the vehicle, and $cnt_{col}$ was increased.
Then, the driving was resumed at a point along the reference path closest to the collision point, as indicated by the yellow line in Fig. \ref{fig:map}.
% Then, the driving was resumed at a point closest to the collision point among a reference path indicated in the yellow line in Fig. \ref{fig:map}.
At this point, the vehicle could drive without a collision.
% This point is where the vehicle can drive without a collision.
The length of the reference path was $len_{path}$.
% $len_{path}$ is the length of the reference path.
A lower \textit{collision rate} indicated a safer driving policy.
% A smaller \textit{collision rate} indicates that the driving policy is safer. 
When the rate was 0, the vehicle could reach the finish point without any collision.
% If the rate is 0, the vehicle reaches the finish point without any collision.

\begin{table}[h]
% \begin{center}
\caption{Collision Rate}
\begin{adjustbox}{width=0.99\linewidth}
\label{table:driving_policy}
\begin{tabular}{c|l|c|c|c}
\Xhline{3\arrayrulewidth}
% 231 139 149
\multicolumn{2}{l|}{\multirow{2}{*}{}} & \multicolumn{3}{c}{Parking Lots} \\
\multicolumn{2}{l|}{} & \multicolumn{1}{l|}{\begin{tabular}[c]{@{}l@{}}Fig. \ref{fig:map1}\\ Trained \\\footnotesize{Environment} \end{tabular}} & \multicolumn{1}{l|}{\begin{tabular}[c]{@{}l@{}}Fig. \ref{fig:map2}\\ \textbf{Untrained} \\\footnotesize{Environment} \end{tabular}} & \multicolumn{1}{l}{\begin{tabular}[c]{@{}l@{}}Fig. \ref{fig:map3}\\ \textbf{Untrained} \\\footnotesize{Environment} \end{tabular}} \\ \hline \hline
\multirow{2}{*}{\begin{tabular}[c]{@{}c@{}}Imitation\\ Learning\end{tabular}} 
 & \multirow{2}{*}{\begin{tabular}[c]{@{}l@{}}\textbf{\textit{DAgger}}\\ \scriptsize{(Proposed)}\end{tabular}} & \multirow{2}{*}{\textbf{0 (0)}} & \multirow{2}{*}{\textbf{0 (0)}} & \multirow{2}{*}{\textbf{0 (0)}} \\
 &  &  &  &  \\ \hline
\multirow{4}{*}{\begin{tabular}[c]{@{}c@{}}Model-based\\Motion\\Planning\end{tabular}} 
& \multirow{2}{*}{\textit{Tentacle}} & \multirow{2}{*}{1.12 (0.95)} & \multirow{2}{*}{1.87 (2.01)} & \multirow{2}{*}{1.47 (1.15)} \\
 &  &  &  &  \\ \cline{2-5} 
%  3+3+2+3+2=13/231/5*100=1.12 (2+2+3+2+2=11/231/5*100=0.95) / 3+3+2+3+2=13/139/5*100=1.87 (4+3+2+2+3=14/139/5*100=2.01) / 2+2+2+3+2=11/149/5*100=1.47 (1+2+2+2+1=8/149/5*100=1.15)
& \multirow{2}{*}{\textit{VVF}} & \multirow{2}{*}{1.38 (1.29)} & \multirow{2}{*}{2.01 (2.15)} & \multirow{2}{*}{1.07 (0.93)} \\
 &  &  &  &  \\
\Xhline{3\arrayrulewidth}
%  4+3+3+3+3=16/231/5*100=1.38 (3+3+3+3+3=15/231/5*100=1.29) / 3+3+3+3+2=14/139/5*100=2.01 (3+3+4+3+2=15/139/5*100=2.15) / 2+1+1+2+2=8/149/5*100=1.07 (2+1+1+2+1=7/149/5*100=0.93)
%  
\end{tabular}
\end{adjustbox}
% \end{center}
{\\ \footnotesize *\textit{Note}.
The values represent the average \textit{collision rate} according to (\ref{eq:collision_rate}) over five trials.
The values in parentheses indicate additional test results where the vehicle turned from the finish point and drove to the start point.
}
\end{table}

Table \ref{table:driving_policy} presents the test results for the \textit{collision rate} at the three parking lots over five trials.
% Table \ref{table:driving_policy} shows the \textit{collision rate} tested in three parking lots with five trials.
In the experiment, each algorithm was used to travel a distance of 5180 m.
% In the experiment, the distance traveled by each algorithm is 5180 m.
The vehicle using \textit{DAgger} method did not encounter any collisions. 
Even in the untrained parking lot with obstacles of different sizes and shapes, the vehicle drove without any collisions.
This result demonstrates that the proposed method has generality.
% which shows that the proposed method is generally effective.
The \textit{tentacle} and \textit{VVF} algorithms resulted in averages of 1.428 and 1.471 collisions per 100 m, respectively.
Several unsafe or near-collision (near-collision) situations occurred with the \textit{tentacle} and \textit{VVF} algorithms as described in the next subsections.

\subsubsection{Safe Distance Range Ratio} 
\begin{figure}[h]
\centering
	\includegraphics[width = 1.0\linewidth]{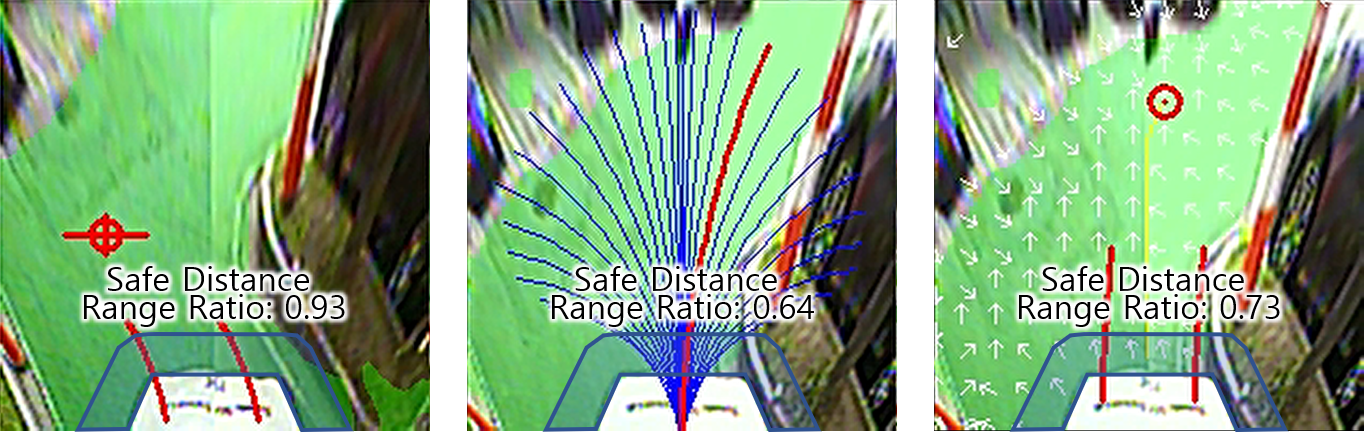}
    \caption{Results of Safe Distance Range Ratio; The blue area is the safe distance range, and its ratio is a measure of how much drivable space (green) exist within the blue area.}
    \label{fig:safe_distance}
\end{figure}

Additionally, in order to evaluate a collision safety with obstacles, a ratio of the drivable space within 1.0 m range from the end of ego vehicle's bumper was measured:
\begin{equation}
    \textit{safe ratio} = \frac{N_{dri}}{N_{ran}},
\label{eq:safe_distance_range_ratio}
\end{equation}
where $N_{dri}$ is the number of pixels for the drivable area among $N_{ran}$.
$N_{ran}$ is the number of pixels around 1.0 m range from the end of ego vehicle's bumper, which is indicated in blue range in Fig. \ref{fig:safe_distance}.
By measuring this ratio, it can be possible to measure how safely the vehicle can maintain a safe distance from obstacles on average.
% By measuring this ratio, you can measure how far away you are from obstacles that are close to the vehicle while driving.
This range and ratio is shown in Fig. \ref{fig:safe_distance} and indicated in Table \ref{table:safe_distance}.
The proposed algorithm \textit{DAgger} has the highest safe distance range ratio.

\begin{table}[h]
% \begin{center}
\caption {Safe Distance Range Ratio}
\begin{adjustbox}{width=1.0\linewidth}
\label{table:safe_distance}
\begin{tabular}{c|l|c|c|c}
\Xhline{3\arrayrulewidth}
\multicolumn{2}{l|}{\multirow{2}{*}{}} & \multicolumn{3}{c}{Parking Lots} \\
\multicolumn{2}{l|}{} & \multicolumn{1}{l|}{\begin{tabular}[c]{@{}l@{}}Fig. \ref{fig:map1}\\ Trained \\\footnotesize{Environment} \end{tabular}} & \multicolumn{1}{l|}{\begin{tabular}[c]{@{}l@{}}Fig. \ref{fig:map2}\\ \textbf{Untrained} \\\footnotesize{Environment} \end{tabular}} & \multicolumn{1}{l}{\begin{tabular}[c]{@{}l@{}}Fig. \ref{fig:map3}\\ \textbf{Untrained} \\\footnotesize{Environment} \end{tabular}} \\ \hline \hline
\multirow{2}{*}{\begin{tabular}[c]{@{}c@{}}Imitation\\ Learning\end{tabular}} 
 & \multirow{2}{*}{\begin{tabular}[c]{@{}l@{}}\textbf{\textit{DAgger}}\\ \scriptsize{(Proposed)}\end{tabular}} & \multirow{2}{*}{\textbf{0.83}} & \multirow{2}{*}{\textbf{0.72}} & \multirow{2}{*}{\textbf{0.91}} \\
 &  &  &  &  \\ \hline
\multirow{4}{*}{\begin{tabular}[c]{@{}c@{}}Model-based\\Motion\\Planning\end{tabular}} 
& \multirow{2}{*}{\textit{Tentacle}} & \multirow{2}{*}{0.69} & \multirow{2}{*}{0.63} & \multirow{2}{*}{0.81} \\
 &  &  &  &  \\ \cline{2-5} 
& \multirow{2}{*}{\textit{VVF}} & \multirow{2}{*}{0.71} & \multirow{2}{*}{0.64} & \multirow{2}{*}{0.85} \\
 &  &  &  &  \\
\Xhline{3\arrayrulewidth}
\end{tabular}
\end{adjustbox}
% \end{center}
{\\ \footnotesize *\textit{Note}. The values represent the average \textit{safe distance range ratio} over five trials.}
\end{table}

\begin{figure*}[t]
\centering
	\includegraphics[width = 1.0\linewidth]{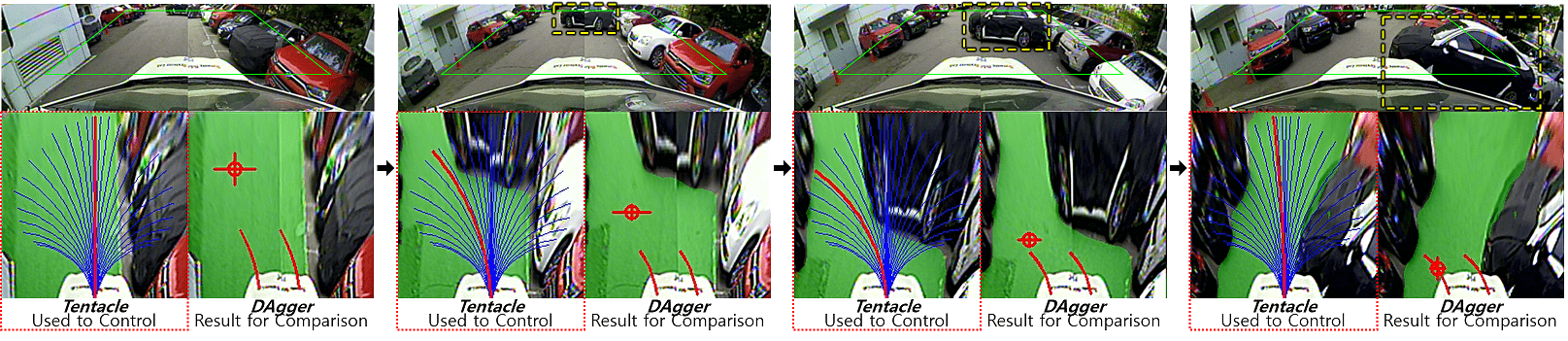}
    \caption{
        The vehicle using \textit{tentacle} did not drive in the middle of the drivable space and did not avoid obstacles safely.
        % The results of \textit{DAgger} are shown together for comparison to its effectiveness.
        The blue lines in the \textit{tentacle} image represent the candidate paths.
        The red line represents the selecte path to track.
        }
    \label{fig:tentacle}
\end{figure*}
% \subsection{ \textit{tentacle} and \textit{VVF} Algorithms}
\subsection{Limitations of \textit{Tentacle} Algorithm}
In the \textit{tentacle} algorithm test, the vehicle drove near the boundary between the drivable and the non-drivable spaces rather than the center of the drivable space after avoiding obstacles or escaping the corner, which is shown in the leftmost image of Fig. \ref{fig:tentacle}.
This is because the \textit{tentacle} algorithm selected the most forward-facing candidate path with no obstacle among the candidate paths.
Then, the vehicle drove at the minimum distance from side obstacles, which increased the possibility of collision.
% If global information is used, the vehicle can be directed to the global path in the middle of the drivable space.
% 
In the same situation, \textit{DAgger} tried to direct the vehicle toward the center of the drivable space.
% In the same situation, unlike \textit{Tentacle}, the proposed method \textit{DAgger} tried to toward the center of the drivable space.
This is because, when the training dataset was collected, experts kept the distance between the vehicle and obstacles as large as possible by considering the overall pattern of the occupancy grid map.
% This is because when the expert collected the training dataset, the vehicle kept the distance from the obstacles as large as possible by considering the overall pattern of the occupancy grid map.

The second to fourth images in Fig. \ref{fig:tentacle} show that, when the vehicle was driving on the side of the drivable space and there was an obstacle in front, the vehicle was unable to avoid the obstacle because of the lack of sufficient space to avoid it.
% Moreover, as shown in the second to fourth figures in Fig. \ref{fig:tentacle}, when the vehicle was driving on the side of the drivable space, and there was an obstacle in front of it, it was unable to avoid the obstacle.
% It is because there was no sufficient space to avoid it.
In other situations, even when a vehicle drove along the center of the drivable space and avoided obstacles, it did not avoid the obstacle with sufficient clearance.
% In other situations, even if the vehicle drove in the center of the drivable space, and there was an obstacle in front of it, the risk to collide still existed.
These are because the \textit{tentacle} algorithm chose the candidate path with the least spacing to avoid obstacles.
% This is because \textit{Tentacle} chose a candidate path with the least spacing to avoid obstacles.
In contrast, \textit{DAgger} tried to avoid obstacles with sufficient safe distance in advance.

\begin{figure}[h]
    \begin{center}
        \subfigure[]{\includegraphics[width=0.99\linewidth]{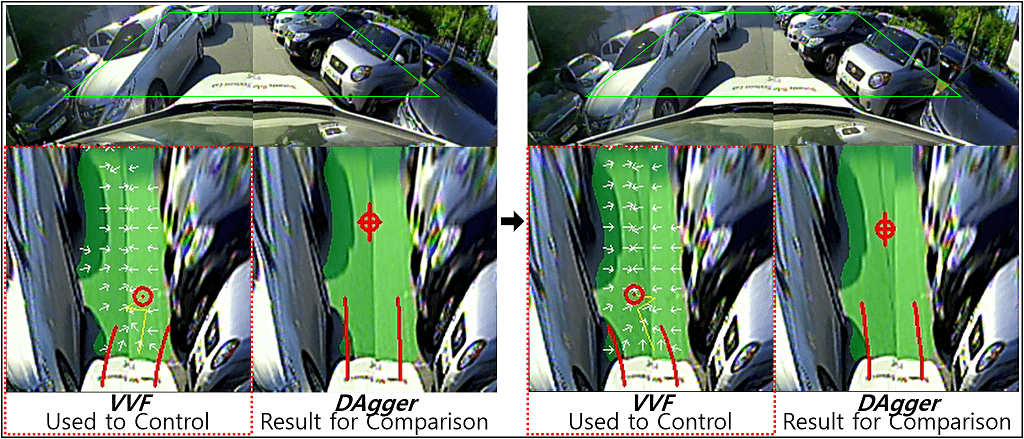}\label{fig:vvf_oscillation}}
        \subfigure[]{\includegraphics[width=0.99\linewidth]{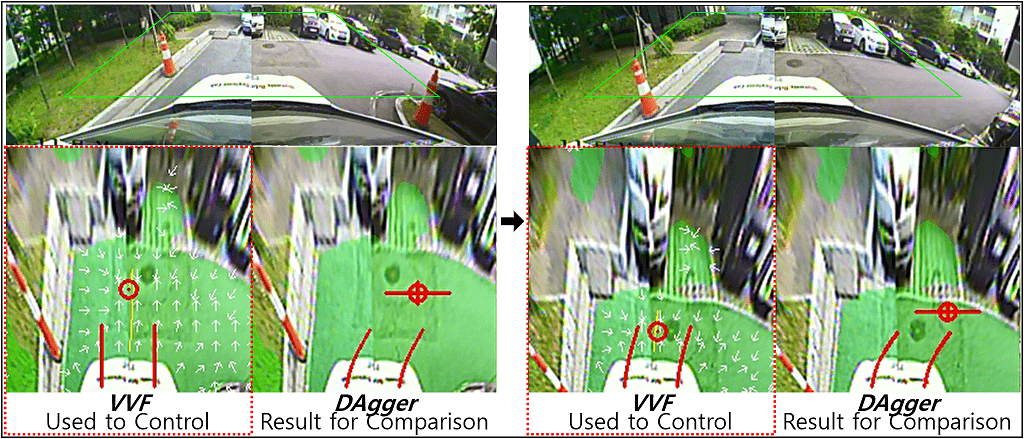}\label{fig:vvf_corner}}
    % 
    % \begin{subfigure}
    %     {\includegraphics[width=0.95\linewidth]{image/8_a VVF_oscillation.png}\label{fig:vvf_oscillation}}
    %     % \sidesubfloat \small{(a)}
    % \end{subfigure}
    % \begin{subfigure}
    %     {\includegraphics[width=0.95\linewidth]{image/8_b VVF_corner.png}\label{fig:vvf_corner}}
    %     % \sidesubfloat \small{(b)}
    % \end{subfigure}
    % 
\end{center}
\caption{
    Problems with the \textit{VVF} algorithm;
    The white arrows represent the field direction.
    (a) Oscillation in a narrow drivable space.
    (b) The vehicle could not enter the side of the drivable space in advance at a right-angled corner.
    \textit{DAgger} did not encounter any problems in situations (a) and (b).
}
\label{fig:vvf}
\end{figure}

\subsection{Limitations of \textit{VVF} Algorithm}
In the \textit{VVF} test, an oscillation problem occurred in narrow drivable spaces where the vehicle frequently turned left and right as shown in Fig. \ref{fig:vvf_oscillation}.
% In the \textit{VVF} test, in a narrow drivable space, there was the oscillation problem that the vehicle frequently turns left and right. 
% This is shown in Fig. \ref{fig:vvf_oscillation}.
In such spaces, because only a repulsive field was applied, the magnitudes of the fields from two obstacles were almost the same, but the directions were opposite.
Thus, the position of the look-ahead point changed frequently in the opposite directions.
This problem may be reduced by decreasing the gain and the range of the repulsive force.
However, the probability of collision would be increased in other situations, especially where the curvature changed significantly.
With \textit{DAgger}, the vehicle drove stably without oscillation by imitating the expert who drove toward the middle of the drivable space even in narrow spaces.
% The vehicle using \textit{DAgger} drove stably without oscillation by imitating the expert driving that drives toward the middle of the drivable space even in narrow spaces.

As shown in Fig. \ref{fig:vvf_corner}, with \textit{VVF}, the vehicle could not enter the drivable space when the curvature changed rapidly, such as right-angled corners.
This problem may be addressed with the global information, where the goal point would be used as an attractive field.
In contrast, this problem did not occur with \textit{DAgger}.
% Unlike \textit{VVF}, when using \textit{DAgger}, this problem did not occur.
This is because, when the training set for \textit{DAgger} was collected in this situation, the expert selected a look-ahead point for which the vehicle could drive the furthest without causing a collision.
% This is because, when collecting the \textit{DAgger} training dataset in this situation, the expert selected the look-ahead point where the vehicle could drive toward the furthest drivable space from the vehicle without causing collision.

\begin{figure}[h]
    \begin{center}
        \subfigure[]{\includegraphics[width=0.99\linewidth]{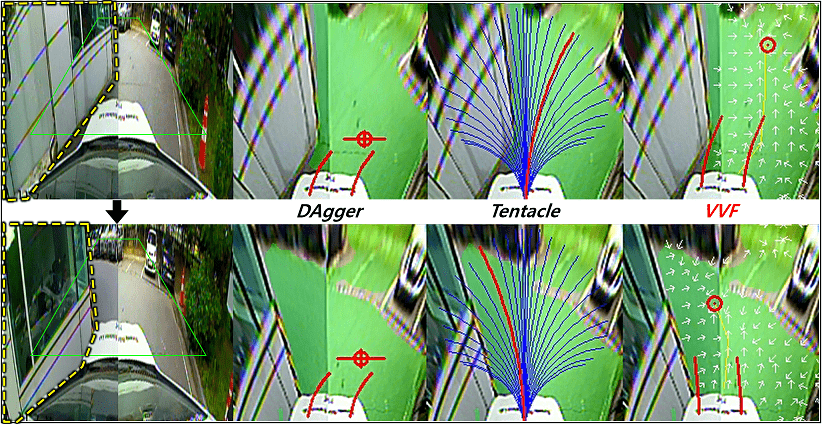}\label{fig:corner1}}
        \subfigure[]{\includegraphics[width=0.99\linewidth]{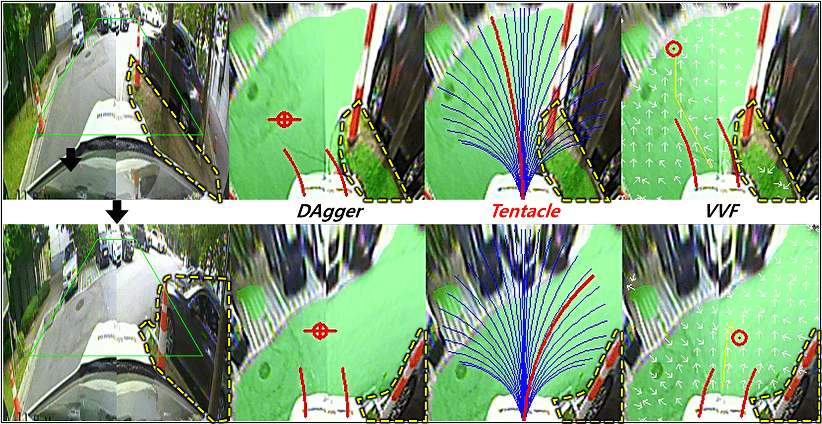}\label{fig:corner2}}
    \end{center}
    % 
    % \begin{subfigure}
    %     {\includegraphics[width=0.95\linewidth]{image/9_a corner.png}\label{fig:corner1}}
    %     % \sidesubfloat \small{(a)}
    % \end{subfigure}
    % \begin{subfigure}
    %     {\includegraphics[width=0.95\linewidth]{image/9_b corner2.png}\label{fig:corner2}}
    %     % \sidesubfloat \small{(b)}
    % \end{subfigure}
\caption{
Problems for driving in narrow drivable space with large curvature changes:
(a) \textit{VVF},
(b) \textit{Tentacle}.
% The driving results for either algorithm were similar.
}
\label{fig:corner}
\end{figure}

\subsection{Limitations of Both \textit{Tentacle} and \textit{VVF} Algorithms}
Figure \ref{fig:corner} shows the problems of the \textit{VVF} and \textit{tentacle} algorithms when the curvature and width of the drivable space changed more than the space where the vehicle was currently driving.
The vehicle headed into the drivable space on the side of the adjacent obstacle before sufficiently avoiding it.
For the \textit{tentacle} algorithm, this is because it selected the path with the fewest obstacles among the candidate paths.
% The reason is that \textit{Tentacle} selected a path with the fewest obstacles just around the candidate paths.
The candidate path set according to the desired velocity (2.2 m/s) was limited in its ability to handle these situations.
% The candidate path set according to the desired velocity (2.2 m/s) had limitations in handling these situations.
For the \textit{VVF} algorithm, the generated field could not sufficiently consider the nearest obstacles.
% The field created in \textit{VVF} could not sufficiently consider the nearest obstacles.
To address this problem, the range of the repulsive field should be increased.
Meanwhile, \textit{DAgger} tried to dodge the nearest obstacle until \textit{DAgger} successfully avoided it because it learned the pattern of preferentially avoiding the nearest obstacles from experts.
% Unlike them, \textit{DAgger} tried to avoid the nearest obstacle until it had been avoided.
% It is because \textit{DAgger} learned a pattern of preferentially avoiding the nearest obstacles.

\begin{figure}[!t]
    % \begin{center}
    %     \subfigure[]{\includegraphics[width=0.95\linewidth]{image/10_a noisy.png}\label{fig:noisy1}}
    %     \subfigure[]{\includegraphics[width=0.95\linewidth]{image/10_b noisy.png}\label{fig:noisy2}}
    %     \subfigure[]{\includegraphics[width=0.95\linewidth]{image/10_c noisy.png}\label{fig:noisy3}}
    %     \subfigure[]{\includegraphics[width=0.95\linewidth]{image/10_d noisy.png}\label{fig:noisy4}}
    %     \subfigure[]{\includegraphics[width=0.95\linewidth]{image/10_e noisy.png}\label{fig:noisy5}}
    %     \subfigure[]{\includegraphics[width=0.95\linewidth]{image/10_f noisy.png}\label{fig:noisy_tentacle}}
    %     \subfigure[]{\includegraphics[width=0.95\linewidth]{image/10_g noisy.png}\label{fig:noisy_vvf}}
%     \end{center}
% % 
    \begin{center}
    \begin{subfigure}
        {\includegraphics[width=0.99\linewidth]{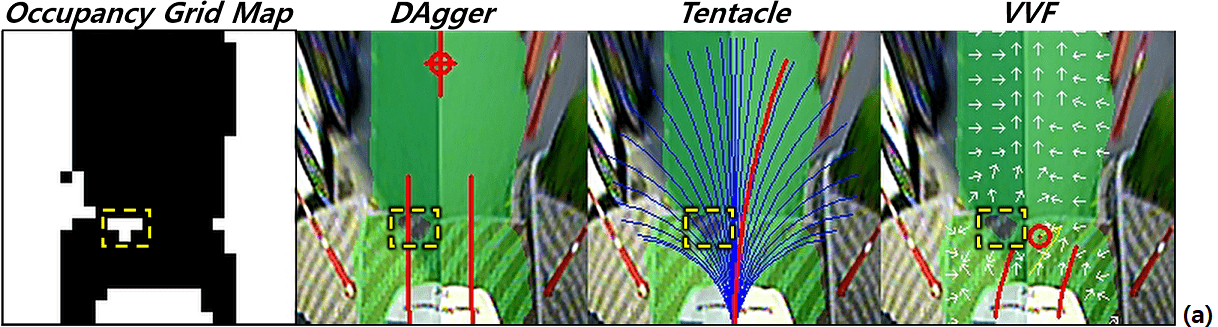}\label{fig:noisy1}}
        % \sidesubfloat \small{(a)}
    \end{subfigure}
    \begin{subfigure}
        {\includegraphics[width=0.99\linewidth]{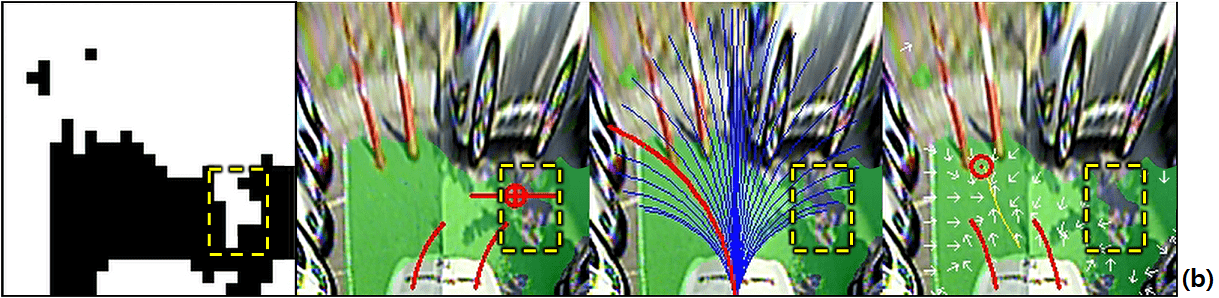}\label{fig:noisy2}}
        % \sidesubfloat \small{(b)}
    \end{subfigure}    
    \begin{subfigure}
        {\includegraphics[width=0.99\linewidth]{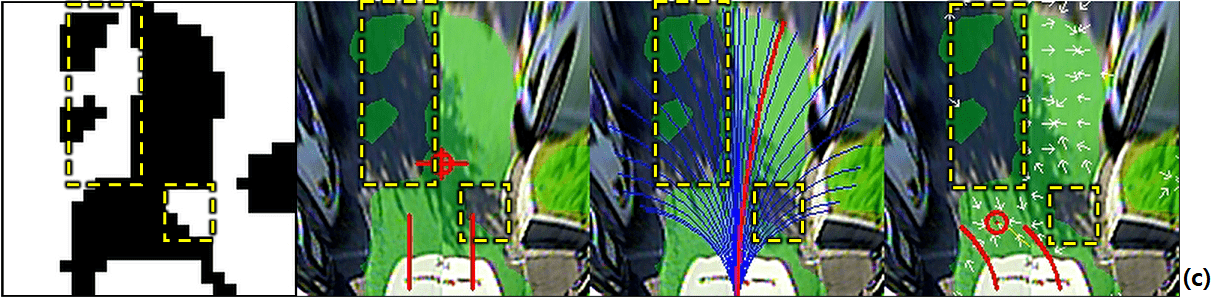}\label{fig:noisy3}}
        % \sidesubfloat \small{(c)}
    \end{subfigure}
    \begin{subfigure}
        {\includegraphics[width=0.99\linewidth]{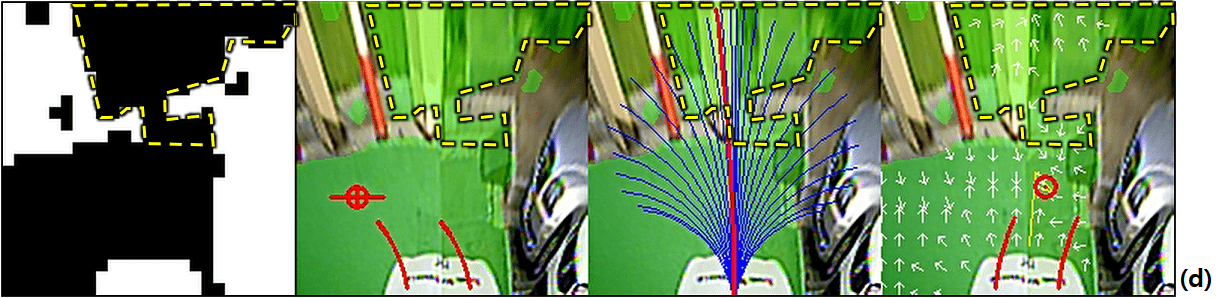}\label{fig:noisy4}}
        % \sidesubfloat \small{(d)}
    \end{subfigure}
    \begin{subfigure}
        {\includegraphics[width=0.99\linewidth]{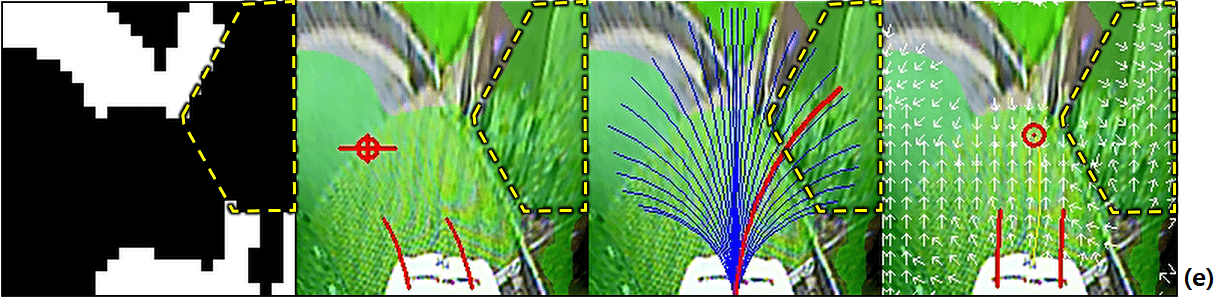}\label{fig:noisy5}}
        % \sidesubfloat \small{(e)}
    \end{subfigure}
    \begin{subfigure}
        {\includegraphics[width=0.99\linewidth]{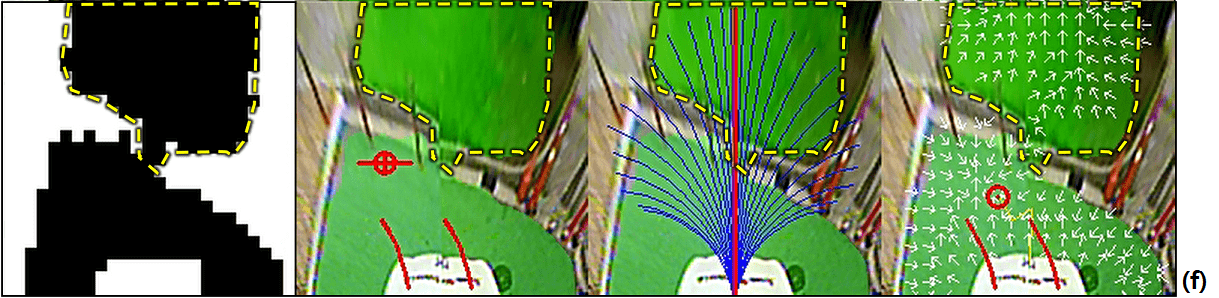}\label{fig:noisy_tentacle}}
        % \sidesubfloat \small{(f)}
    \end{subfigure}
    \begin{subfigure}
        {\includegraphics[width=0.99\linewidth]{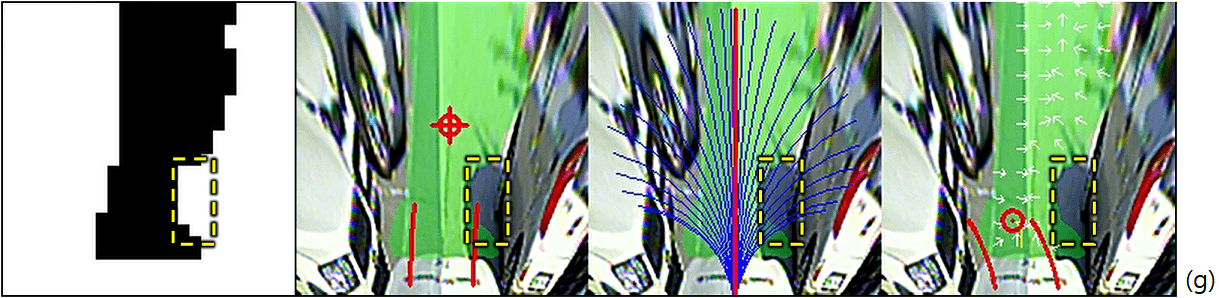}\label{fig:noisy_vvf}}
        % \sidesubfloat \small{(g)}
    \end{subfigure}
    \end{center}
\caption{
Driving results with \textit{DAgger} when the occupancy grid map contained noise.
\textit{DAgger} did not encounter any problems in this situation.
However, the vehicle could not drive smoothly or headed towards obstacles with \textit{Tentacle} and \textit{VVF}.
(a) Noise from misrecognition; (b), (c), and (g) Noise by shadow; (d), (e), and (f) Noise at the road boundary (misrecognition).
}
\label{fig:noisy}
\end{figure}

\subsection{Driving Results on Noisy Occupancy Grid Map}
% noisy
The occupancy grid map was not recognized accurately in complex and shadowy environments (i.e., noisy state) because the learning data for such situations were insufficient to train the perception network.
Data with the noisy state were contained in training data, so the trained network could learn some patterns for the noise and deal with the noisy state.
As can be seen from the experiment in Fig. \ref{fig:noisy}, a vehicle could drive without collision, even though there were noise in one trained environment (see Fig. \ref{fig:noisy1}, \ref{fig:noisy2}, and \ref{fig:noisy3}) and two untrained environments (see Fig. \ref{fig:noisy4}, \ref{fig:noisy5}, \ref{fig:noisy_tentacle}, and \ref{fig:noisy_vvf}).
% It is because when training the perception deep neural network, learning data with complex situations and shadow were insufficient \cite{varma2019idd}.
However, the \textit{Tentacle} and \textit{VVF} algorithms encountered several problems.
% Figure \ref{fig:noisy} compares the driving results with \textit{DAgger} to those with the \textit{Tentacle} and \textit{VVF} algorithms.
%% The driving results of \textit{Tentacle} and \textit{VVF} algorithms were same as those of \textit{DAgger}.
%% Besides, the results of executing \textit{Tentacle} and \textit{VVF} are shown together.
%% The results driving with \textit{tentacle} and \textit{VVF} were almost identical to the results of each algorithm when driving with \textit{DAgger} as shown in the figure \ref {fig:noisy}.

As shown in Fig. \ref{fig:noisy1}, the boundary between the speed bump and road was erroneously recognized as non-drivable space (i.e., noise).
In this situation, \textit{DAgger} was not affected by the noise because it trained the driving pattern from the overall shape of the state.
However, the vehicle drove unstably with the \textit{Tentacle} and \textit{VVF} algorithms to avoid the misrecognized non-drivable space.
 
Figures \ref{fig:noisy2} and \ref{fig:noisy3} show situations where noise was caused by shadows.
With \textit{DAgger}, the vehicle drove towards the drivable space with fewer oscillations than \textit{tentacle} and \textit{VVF}.
This is because the \textit{DAgger} training dataset contained similar situations, where the expert selected an action without being affected by the noise.
% The reason is that similar situations were contained in the \textit{DAgger} training dataset, and in these situations the expert selected the action without being affected by the noise.
With the \textit{tentacle} and \textit{VVF} algorithms, however, the vehicle in the Fig. \ref{fig:noisy2} situation avoided the shadows and then drove toward the largest drivable space blocked by obstacles, so it became unable to drive any further.
% However, in Fig. \ref{fig:noisy2} situation, The vehicle using \textit{Tentacle} and \textit{VVF} avoided the shadow and then drove toward the largest drivable space blocked by obstacles, so it could not drive anymore.
These algorithms also had larger oscillation problems than \textit{DAgger} especially in the Fig. \ref{fig:noisy3} situation.

Figures \ref{fig:noisy4}, \ref{fig:noisy5}, and \ref{fig:noisy_tentacle} present situations in which a non-drivable space was recognized as drivable space.
In detail, not only the non-drivable space at the curb (i.e., boundary of the drivable space) but also the space behind the curb was recognized as drivable space.
% In detail, it was recognized that not only the non-drivable space at the kerb (boundary of the drivable space), but also the rear space of this space was recognized as the drivable space.
With \textit{DAgger}, the vehicle tried to drive toward the largest drivable space, except for behind the curb.
% The vehicle with \textit{DAgger} tried to drive toward the largest drivable space, except behind the kerb.
However, \textit{tentacle} was influenced by the noise at the curb, which it detected to be drivable space.
So, the vehicle was headed to the curb.
\textit{VVF} was less affected than the \textit{tentacle} algorithm, but the vehicle was unable to drive toward the largest drivable space (see Figs. \ref{fig:noisy4} and \ref{fig:noisy5}).

As shown in Fig. \ref{fig:noisy_vvf}, the vehicle with the \textit{VVF} algorithm took actions to avoid the noise caused by a shadow next to the obstacle when passing through a narrow space.
For the same situation, \textit{DAgger} and the \textit{tentacle} algorithm did not respond sensitively, and no problem occurred.

\subsection{Analyses of Stability and Time-delay}
Although the stability cannot be theoretically proven in imitation learning, it has been experimentally confirmed that there is no problem in the parking lot environment through sufficient training with \textit{DAggar}.
For a dynamic obstacle, if the speed is lower than about 10 km/h, it can be treated similarly to a static obstacle, and there was no problem in the actual experiment.

% \item Nonlinear Characteristics:
% The proposed method can be regarded as a non-linear system.
% This is because the deep neural networks for perception and driving used softmax, Sigmoid, and Relu activation functions which are the non-linear function.
The time-delay problem did not occur because all parts were calculated within a defined control period of the proposed algorithm, 50 ms.
The perception and driving networks calculated each outputs within 20 ms and 10 ms.
Besides, obtaining control commands to reach the look-ahead point were calculated in 1 ms, and the vehicle responded to these commands within 30 ms.
%%%%%%%%%%%%%%%%%%%%%%%%%%%%%%%%%%%%%%%%
\section{Conclusion}
\label{sec:conclusion}
In this study, an autonomous driving method using vision-based occupancy grid map and imitation learning is proposed to deal with unstructured environments such as parking lots.
% This study proposes a method of autonomous driving in unstructured environments such as parking lot using vision-based occupancy grid map and imitation learning.
With the proposed method, the vehicle can drive toward the drivable space while avoiding obstacles reactively in real-time without using a global map and localization.
Besides, it does not need to model the driving policy and tune model-parameters of the policy.
% The vehicle using the proposed method can drive toward the drivable space while avoiding obstacles reactively in real-time without using global map and localization data.
% The vision data is passed through the deep neural network based on U-Net to obtain the occupancy grid map.
% The vision data is passed through the U-net-based deep neural network, and the occupancy grid map is obtained.
The occupancy grid map obtained by the U-net-based deep neural network is used as an input for imitation learning, where the driving patterns of experts in various and complex environments are learned.
% Imitation learning takes this map as input and is used to imitate the expert driving pattern in various and complex environments.
In experiments, a real autonomous vehicle was trained with \textit{DAgger} in one parking lot and tested in three parking lots (1036 m) without intersections five times each (totaling 5180 m).
% It was trained in one parking lot (460 m) with a real autonomous vehicle using the \textit{DAgger} algorithm and tested in three parking lots (1,036 m) with five times (totaling 5,180 m) without intersections.

With \textit{DAgger}, the vehicle could drive more smoothly and safely than with the \textit{tentacle} and \textit{VVF} algorithms in environments where the width and curvature of the drivable space varied significantly.
% \textcolor{red}{Especially, \textit{DAgger} was more robust when the occupancy grid map was noisy due to a shadow or a misrecognition problem}.
Especially, \textit{DAgger} was more robust when the occupancy grid map was not accurately perceived or was noisy due to a shadow.
With regard to the \textit{collision rate}, \textit{DAgger} did not cause any collision, but the \textit{tentacle} and \textit{VVF} algorithms caused 1.42 and 1.47 collisions per 100 m, respectively.
% As the result of measuring the number of times close to collision per 100 m, \textit{DAgger} did not cause any collision, but \textit{Tentacle} and \textit{VVF} occurred 1.42 and 1.47 times, respectively.
This is because the \textit{tentacle} and \textit{VVF} algorithms require different parameters to accommodate different complex situations.
In contrast, \textit{DAgger} trains the deep neural network with numerous weight parameters using expert driving data for these situations.
Future work will focus on developing the proposed method to environments with intersections and dynamic obstacles.

\bibliographystyle{IEEEtran}
\bibliography{mybibfile}

\begin{thebibliography}{10}
\providecommand{\url}[1]{#1}
\csname url@samestyle\endcsname
\providecommand{\newblock}{\relax}
\providecommand{\bibinfo}[2]{#2}
\providecommand{\BIBentrySTDinterwordspacing}{\spaceskip=0pt\relax}
\providecommand{\BIBentryALTinterwordstretchfactor}{4}
\providecommand{\BIBentryALTinterwordspacing}{\spaceskip=\fontdimen2\font plus
\BIBentryALTinterwordstretchfactor\fontdimen3\font minus
  \fontdimen4\font\relax}
\providecommand{\BIBforeignlanguage}[2]{{%
\expandafter\ifx\csname l@#1\endcsname\relax
\typeout{** WARNING: IEEEtran.bst: No hyphenation pattern has been}%
\typeout{** loaded for the language `#1'. Using the pattern for}%
\typeout{** the default language instead.}%
\else
\language=\csname l@#1\endcsname
\fi
#2}}
\providecommand{\BIBdecl}{\relax}
\BIBdecl

\bibitem{banzhaf2017future}
H.~Banzhaf, D.~Nienh{\"u}ser, S.~Knoop, and J.~M. Z{\"o}llner, ``The future of
  parking: A survey on automated valet parking with an outlook on high density
  parking,'' in \emph{2017 IEEE Intelligent Vehicles Symposium (IV)}.\hskip 1em
  plus 0.5em minus 0.4em\relax IEEE, 2017, pp. 1827--1834.

\bibitem{paden2016survey}
B.~Paden, M.~{\v{C}}{\'a}p, S.~Z. Yong, D.~Yershov, and E.~Frazzoli, ``A survey
  of motion planning and control techniques for self-driving urban vehicles,''
  \emph{IEEE Transactions on intelligent vehicles}, vol.~1, no.~1, pp. 33--55,
  2016.

\bibitem{gonzalez2015review}
D.~Gonz{\'a}lez, J.~P{\'e}rez, V.~Milan{\'e}s, and F.~Nashashibi, ``A review of
  motion planning techniques for automated vehicles,'' \emph{IEEE Transactions
  on Intelligent Transportation Systems}, vol.~17, no.~4, pp. 1135--1145, 2015.

\bibitem{shin2021model}
J.~Shin, D.~Kwak, and K.~Kwak, ``Model predictive path planning for an
  autonomous ground vehicle in rough terrain,'' \emph{International Journal of
  Control, Automation and Systems}, vol.~19, no.~6, pp. 2224--2237, 2021.

\bibitem{mousavi2017new}
M.~A. Mousavi, B.~Moshiri, and Z.~Heshmati, ``A new predictive motion control
  of a planar vehicle under uncertainty via convex optimization,''
  \emph{International Journal of Control, Automation and Systems}, vol.~15,
  no.~1, pp. 129--137, 2017.

\bibitem{al2020voronoi}
M.~R.~H. Al-Dahhan and K.~W. Schmidt, ``Voronoi boundary visibility for
  efficient path planning,'' \emph{IEEE Access}, vol.~8, pp.
  134\,764--134\,781, 2020.

\bibitem{niu2019voronoi}
H.~Niu, A.~Savvaris, A.~Tsourdos, and Z.~Ji, ``Voronoi-visibility roadmap-based
  path planning algorithm for unmanned surface vehicles,'' \emph{The Journal of
  Navigation}, vol.~72, no.~4, pp. 850--874, 2019.

\bibitem{mohanta2019knowledge}
J.~C. Mohanta and A.~Keshari, ``A knowledge based fuzzy-probabilistic roadmap
  method for mobile robot navigation,'' \emph{Applied Soft Computing}, vol.~79,
  pp. 391--409, 2019.

\bibitem{dijkstra1959note}
E.~W. Dijkstra \emph{et~al.}, ``A note on two problems in connexion with
  graphs,'' \emph{Numerische mathematik}, vol.~1, no.~1, pp. 269--271, 1959.

\bibitem{hart1968formal}
P.~E. Hart, N.~J. Nilsson, and B.~Raphael, ``A formal basis for the heuristic
  determination of minimum cost paths,'' \emph{IEEE transactions on Systems
  Science and Cybernetics}, vol.~4, no.~2, pp. 100--107, 1968.

\bibitem{shin2016desired}
S.~Shin, J.~Ahn, and J.~Park, ``Desired orientation rrt (do-rrt) for autonomous
  vehicle in narrow cluttered spaces,'' in \emph{2016 IEEE/RSJ International
  Conference on Intelligent Robots and Systems (IROS)}.\hskip 1em plus 0.5em
  minus 0.4em\relax IEEE, 2016, pp. 4736--4741.

\bibitem{dolgov2009path}
D.~Dolgov, S.~Thrun, M.~Montemerlo, and J.~Diebel, ``Path planning for
  autonomous driving in unknown environments,'' in \emph{Experimental
  Robotics}.\hskip 1em plus 0.5em minus 0.4em\relax Springer, 2009, pp. 55--64.

\bibitem{likhachev2008anytime}
M.~Likhachev, D.~Ferguson, G.~Gordon, A.~Stentz, and S.~Thrun, ``Anytime search
  in dynamic graphs,'' \emph{Artificial Intelligence}, vol. 172, no.~14, pp.
  1613--1643, 2008.

\bibitem{hoy2015algorithms}
M.~Hoy, A.~S. Matveev, and A.~V. Savkin, ``Algorithms for collision-free
  navigation of mobile robots in complex cluttered environments: a survey,''
  \emph{Robotica}, vol.~33, no.~3, pp. 463--497, 2015.

\bibitem{missura2019predictive}
M.~Missura and M.~Bennewitz, ``Predictive collision avoidance for the dynamic
  window approach,'' in \emph{2019 International Conference on Robotics and
  Automation (ICRA)}.\hskip 1em plus 0.5em minus 0.4em\relax IEEE, 2019, pp.
  8620--8626.

\bibitem{lopez2019new}
J.~L{\'o}pez, C.~Otero, R.~Sanz, E.~Paz, E.~Molinos, and R.~Barea, ``A new
  approach to local navigation for autonomous driving vehicles based on the
  curvature velocity method,'' in \emph{2019 International Conference on
  Robotics and Automation (ICRA)}.\hskip 1em plus 0.5em minus 0.4em\relax IEEE,
  2019, pp. 1751--1757.

\bibitem{mouhagir2019evidential}
H.~Mouhagir, R.~Talj, V.~Cherfaoui, F.~Aioun, and F.~Guillemard,
  ``Evidential-based approach for trajectory planning with tentacles, for
  autonomous vehicles,'' \emph{IEEE Transactions on Intelligent Transportation
  Systems}, vol.~21, no.~8, pp. 3485--3496, 2019.

\bibitem{olunloyo2009autonomous}
V.~Olunloyo and M.~Ayomoh, ``Autonomous mobile robot navigation using hybrid
  virtual force field concept,'' \emph{European Journal of Scientific
  Research}, vol.~31, no.~2, pp. 204--228, 2009.

\bibitem{ge2000new}
S.~S. Ge and Y.~J. Cui, ``New potential functions for mobile robot path
  planning,'' \emph{IEEE Transactions on robotics and automation}, vol.~16,
  no.~5, pp. 615--620, 2000.

\bibitem{wang2015driving}
J.~Wang, J.~Wu, and Y.~Li, ``The driving safety field based on
  driver--vehicle--road interactions,'' \emph{IEEE Transactions on Intelligent
  Transportation Systems}, vol.~16, no.~4, pp. 2203--2214, 2015.

\bibitem{chen2019deep}
J.~Chen, B.~Yuan, and M.~Tomizuka, ``Deep imitation learning for autonomous
  driving in generic urban scenarios with enhanced safety,'' in \emph{2019
  IEEE/RSJ International Conference on Intelligent Robots and Systems
  (IROS)}.\hskip 1em plus 0.5em minus 0.4em\relax IEEE, 2019, pp. 2884--2890.

\bibitem{hawke2020urban}
J.~Hawke, R.~Shen, C.~Gurau, S.~Sharma, D.~Reda, N.~Nikolov, P.~Mazur,
  S.~Micklethwaite, N.~Griffiths, A.~Shah \emph{et~al.}, ``Urban driving with
  conditional imitation learning,'' in \emph{2020 IEEE International Conference
  on Robotics and Automation (ICRA)}.\hskip 1em plus 0.5em minus 0.4em\relax
  IEEE, 2020, pp. 251--257.

\bibitem{youn2021collision}
W.~Youn, H.~Ko, H.~Choi, I.~Choi, J.-H. Baek, and H.~Myung, ``Collision-free
  autonomous navigation of a small uav using low-cost sensors in gps-denied
  environments,'' \emph{International Journal of Control, Automation and
  Systems}, vol.~19, no.~2, pp. 953--968, 2021.

\bibitem{varma2019idd}
G.~Varma, A.~Subramanian, A.~Namboodiri, M.~Chandraker, and C.~Jawahar, ``Idd:
  A dataset for exploring problems of autonomous navigation in unconstrained
  environments,'' in \emph{2019 IEEE Winter Conference on Applications of
  Computer Vision (WACV)}.\hskip 1em plus 0.5em minus 0.4em\relax IEEE, 2019,
  pp. 1743--1751.

\bibitem{lee2021deep}
D.-H. Lee, K.-L. Chen, K.-H. Liou, C.-L. Liu, and J.-L. Liu, ``Deep learning
  and control algorithms of direct perception for autonomous driving,''
  \emph{Applied Intelligence}, vol.~51, no.~1, pp. 237--247, 2021.

\bibitem{hawke2019urban}
J.~Hawke, R.~Shen, C.~Gurau, S.~Sharma, D.~Reda, N.~Nikolov, P.~Mazur,
  S.~Micklethwaite, N.~Griffiths, A.~Shah \emph{et~al.}, ``Urban driving with
  conditional imitation learning,'' \emph{arXiv preprint arXiv:1912.00177},
  2019.

\bibitem{hadfield2017inverse}
D.~Hadfield-Menell, S.~Milli, P.~Abbeel, S.~J. Russell, and A.~Dragan,
  ``Inverse reward design,'' in \emph{Advances in neural information processing
  systems}, 2017, pp. 6765--6774.

\bibitem{bojarski2016end}
M.~Bojarski, D.~Del~Testa, D.~Dworakowski, B.~Firner, B.~Flepp, P.~Goyal, L.~D.
  Jackel, M.~Monfort, U.~Muller, J.~Zhang \emph{et~al.}, ``End to end learning
  for self-driving cars,'' \emph{arXiv preprint arXiv:1604.07316}, 2016.

\bibitem{menda2019ensembledagger}
K.~Menda, K.~Driggs-Campbell, and M.~J. Kochenderfer, ``Ensembledagger: A
  bayesian approach to safe imitation learning,'' in \emph{2019 IEEE/RSJ
  International Conference on Intelligent Robots and Systems, IROS 2019}.\hskip
  1em plus 0.5em minus 0.4em\relax Institute of Electrical and Electronics
  Engineers Inc., 2019, pp. 5041--5048.

\bibitem{teichmann2018multinet}
M.~Teichmann, M.~Weber, M.~Zoellner, R.~Cipolla, and R.~Urtasun, ``Multinet:
  Real-time joint semantic reasoning for autonomous driving,'' in \emph{2018
  IEEE Intelligent Vehicles Symposium (IV)}.\hskip 1em plus 0.5em minus
  0.4em\relax IEEE, 2018, pp. 1013--1020.

\bibitem{simonyan2014very}
K.~Simonyan and A.~Zisserman, ``Very deep convolutional networks for
  large-scale image recognition,'' \emph{arXiv preprint arXiv:1409.1556}, 2014.

\bibitem{ross2011reduction}
S.~Ross, G.~Gordon, and D.~Bagnell, ``A reduction of imitation learning and
  structured prediction to no-regret online learning,'' in \emph{Proceedings of
  the fourteenth international conference on artificial intelligence and
  statistics}, 2011, pp. 627--635.

\bibitem{kendall2017uncertainties}
A.~Kendall and Y.~Gal, ``What uncertainties do we need in bayesian deep
  learning for computer vision?'' in \emph{Advances in neural information
  processing systems}, 2017, pp. 5574--5584.

\bibitem{ahn22accurate}
J.~Ahn, S.~Shin, M.~Kim, and J.~Park, ``Accurate path tracking by adjusting
  look-ahead point in pure pursuit method,'' \emph{International Journal of
  Automotive Technology}, vol.~22, no.~1, pp. 119--129.

\bibitem{kelly2019hg}
M.~Kelly, C.~Sidrane, K.~Driggs-Campbell, and M.~J. Kochenderfer, ``Hg-dagger:
  Interactive imitation learning with human experts,'' in \emph{2019
  International Conference on Robotics and Automation (ICRA)}.\hskip 1em plus
  0.5em minus 0.4em\relax IEEE, 2019, pp. 8077--8083.

\bibitem{byrd2019effect}
J.~Byrd and Z.~Lipton, ``What is the effect of importance weighting in deep
  learning?'' in \emph{International Conference on Machine Learning}.\hskip 1em
  plus 0.5em minus 0.4em\relax PMLR, 2019, pp. 872--881.

\bibitem{Breheret:2017}
A.~Br{\'e}h{\'e}ret, ``{Pixel Annotation Tool},''
  \url{https://github.com/abreheret/PixelAnnotationTool}, 2017.

\end{thebibliography}

\end{document}